\documentclass[sigconf]{acmart}
\AtBeginDocument{%
  \providecommand\BibTeX{{%
    \normalfont B\kern-0.5em{\scshape i\kern-0.25em b}\kern-0.8em\TeX}}}

\copyrightyear{2021}
\acmYear{2021}
\setcopyright{acmcopyright}\acmConference[KDD '21]{Proceedings of the 27th ACM SIGKDD Conference on Knowledge Discovery and Data Mining}{August 14--18, 2021}{Virtual Event, Singapore}
\acmBooktitle{Proceedings of the 27th ACM SIGKDD Conference on Knowledge Discovery and Data Mining (KDD '21), August 14--18, 2021, Virtual Event, Singapore}
\acmPrice{15.00}
\acmDOI{10.1145/3447548.3467420}
\acmISBN{978-1-4503-8332-5/21/08}
\newcommand{\etal}{\textit{et al}.}

\usepackage{enumitem}
\usepackage{algorithmicx,algorithm}
\usepackage[noend]{algpseudocode}
\usepackage{multirow}
\usepackage{subcaption}
\newcommand\blfootnote[1]{%
  \begingroup
  \renewcommand\thefootnote{}\footnote{#1}%
  \addtocounter{footnote}{-1}%
  \endgroup
}

\begin{document}
\fancyhead{} 
\title{Shapley Counterfactual Credits for Multi-Agent Reinforcement Learning}

\author{Jiahui Li$^1$, Kun Kuang$^{1*}$, Baoxiang Wang$^{2,3}$, Furui Liu$^{4*}$, Long Chen$^{1\dagger}$, Fei Wu$^1$, Jun Xiao$^1$}
\affiliation{%
\institution{$^1$DCD Lab, College of Computer Science, Zhejiang University}
\institution{$^2$The Chinese University of Hong Kong, Shenzhen}
\institution{$^3$Shenzhen Institute of Artificial Intelligence and Robotics for Society}
\institution{$^4$Huawei Noah's Ark Lab}
\city{}
\country{}
}
\email{{jiahuil,kunkuang}@zju.edu.cn, bxiangwang@gmail.com,liufurui2@huawei.com}
\email{zjuchenlong@gmail.com, {wufei, junx}@cs.zju.edu.cn}

\begin{abstract}
Centralized Training with Decentralized Execution (CTDE) has been a popular paradigm in cooperative Multi-Agent Reinforcement Learning (MARL) settings and is widely used in many real applications. One of the major challenges in the training process is credit assignment, which aims to deduce the contributions of each agent according to the global rewards.
Existing credit assignment methods focus on either decomposing the joint value function into individual value functions or measuring the impact of local observations and actions on the global value function. 
These approaches lack a thorough consideration of the complicated interactions among multiple agents, leading to an unsuitable assignment of credit and subsequently mediocre results on MARL.
We propose Shapley Counterfactual Credit Assignment, a novel method for explicit credit assignment which accounts for the coalition of agents.
Specifically, Shapley Value and its desired properties are leveraged in deep MARL to credit any combinations of agents, which grants us the capability to estimate the individual credit for each agent. 
Despite this capability, the main technical difficulty lies in the computational complexity of Shapley Value who grows factorially as the number of agents.
We instead utilize an approximation method via Monte Carlo sampling, which reduces the sample complexity while maintaining its effectiveness.
We evaluate our method on StarCraft II benchmarks across different scenarios. 
Our method outperforms existing cooperative MARL algorithms significantly and achieves the state-of-the-art, with especially large margins on tasks with more severe difficulties.
\blfootnote{$^*$ Kun Kuang and Furui Liu are the corresponding authors.}
\blfootnote{$^\dagger$ This work was done when Long Chen was a Ph.D. student at Zhejiang University.}
\end{abstract}

\maketitle
\section{Introduction}
Multi-Agent Systems (MAS) have attracted substantial attention in many sequential decision problems in recent years, such as autonomous vehicle teams~\cite{keviczky2007decentralized,cao2012overview}, robotics~\cite{lillicrap2016continuous,ramchurn2010decentralized}, scene graph generation~\cite{chen2019counterfactual}, and network routing~\cite{ye2015multi}, etc. 
Among the approaches, Multi-Agent Reinforcement Learning (MARL) has grown its popularity with its ability to learn without knowing the world model. 
A classical way in MARL to solve cooperative games is regarding the entire MAS as a single agent and optimize a joint policy according to the joint observations and trajectories~\cite{tan1993multi}. 
With the joint action space of agents growing exponentially as the number of agents and the constraints of partial observability, the classical method faces insurmountable obstacles. 
This promotes the Centralized Training with Decentralized Execution (CTDE)~\cite{oliehoek2008optimal,kraemer2016multi} paradigm, where a central critic is set up to estimate the joint value function, and the agents are trained with global information but executed only based on its local observes and histories. 

The main challenge that restricts the effective CTDE in MARL is credit assignment, which attributes the global reward signals according to the contributions of each agent. Recent studies that attempt to solve this challenge can be roughly divided into two branches. 1) \emph{Implicit methods}~\cite{sunehag2018value,rashid2018qmix,son2019qtran,yang2020qatten}: it treats the central critic and the local agents as an entirety during the training procedure. A decomposition function (usually a neural network) is first set up to map the joint value function to local value functions. The central critic is then learned simultaneously with the decomposition function and the policy. Implicit methods suffer from inadequate decomposition limited by the design of the decomposition function. They also lack the interpretability for the distributed credits~\cite{heuillet2021explainability}.
2) \emph{Explicit methods}~\cite{foerster2018counterfactual,wang2020shapleyq,yang2020q}: it trains the central critic and the local actors separately. In each iteration, the critic is first updated, after which some strategies are leveraged to compute the reward or the value function of each agent explicitly. Such reward signals or value functions are used to guide the training of local agents.
Despite that the explicit methods overcome many shortcomings of the implicit counterpart, one has to algorithmically characterize the individual agent's contribution from the overall success, which can be very hard in the context of subtle coalitions under common goals.
We address this challenge by using a counterfactual method with Shapley Value.
Shapley Value~\cite{shapley1953value} originates from cooperative game theory and is a golden standard to distribute benefits reasonably and fairly by accounting for the contribution of participating players.
By treating the agents in MARL as the players in cooperative games, ideal credit assignment can be obtained up to computing the marginal contribution of Shapley Value. 
Inspired from this, Wang~\etal~\cite{wang2020shapleyq} proposed SQDDPG, which utilized Shapley Value in deterministic policy gradient~\cite{silver2014deterministic,lowe2017multi} to guide the learning of local agents. 
However, the performance of SQDDPG relies highly on the designed framework for estimating the marginal contribution, and this framework is limited by an assumption that the actions of agents are taken sequentially, which is often unrealistic.
These restrictions make SQDDPG perform unsatisfactory in many tasks.
To this end, we extend the explicit methods and propose a novel method that leverages Shapley Value to allocate the credits for agents. 
We achieve it by leveraging a counterfactual method to estimate what would have happened without the participation of a set of agents. 
The quantification of the contribution of a set of agents is then computed as the change of the central critic value by setting their actions to a baseline.
Then the changes of the contributions caused by an agent in different set unions are treated as marginal contributions, and Shapley Value can thus be obtained.
Finally, these unified values play the role of credits in local policies and guide its training procedure.

Nevertheless, the computational complexity of the original Shapley Value grows factorially as the number of players increases.
In many contexts of interest, such as network games, distributed control, and computing economics, this number can be quite large, which makes Shapley Value intractable. 
To alleviate the computational burden, we approximate Shapley Value through Monte Carlo sampling, which maintains the majority of the desired properties of Shapley Value. 
In our approach, the Shapley Value is computed by subsets of collaborators for each agent and is re-sampled at each time step.
Our approach manages to reduce the computational complexity to polynomial in the number of players without much loss of effectiveness of Shapley Value.

Our main contributions can be summarized as follows:
\begin{enumerate}
\item We leverage a counterfactual method with Shapley Value to address the problem of credit assignment in Multi-Agent Reinforcement Learning. The proposed Shapley Counterfactual Credits reasonably and fairly characterize the contributions of each local agent by fully considering their interactions.
\item We adopt a Monte Carlo sampling-based method to approximate Shapley Value and decrease its computational complexity growth from factorial to polynomial, which makes our algorithm viable for large-scale, complicated tasks.
\item Extensive experiments show that our proposed method outperforms existing cooperative MARL algorithms significantly and achieves state-of-the-art performance on StarCraft II benchmarks. The margin is especially large for more difficult tasks.
\end{enumerate}

The rest of this paper is organized as follows.
In \emph{Section~\ref{sec:related work}}, we first briefly reviews all related works. And we introduce the preliminaries, including Dec-POMDPs, Shapley Value, and explicit framework for MARL in \emph{Section~\ref{sec:preliminaries}}. The details of our proposed algorithm for credit assignment are introduced in \emph{Section~\ref{sec:SCC}}. Experimental results and analyses are reported in \emph{Section~\ref{sec:experiments}}.
Finally, we conclude our paper and discuss on future directions in \emph{Section~\ref{sec:conclusion}} .

\section{Related Work} \label{sec:related work}
\subsection{Implicit Credit Assignment}
Most of the implicit methods follow the condition of Individual-Global-Max (IGM), which means the optimal joint actions among the agents are equivalent to the optimal actions of each local agent. VDN~\cite{sunehag2018value} makes a hypothesis of the additivity to decompose the joint Q-function into the sum of individual Q-functions. QMIX~\cite{rashid2018qmix} gets rid of this assumption but adds a restriction of the monotonicity. LICA~\cite{zhou2020learning} promotes QMIX to actor-critic as well as proposes an adaptive entropy regularization. Weighted QMIX adapts a twins network and encourages the underestimated actions to alleviate the risk of suboptimal results. QTRAN~\cite{son2019qtran} avoids the limitations of VDN and QMIX by introducing two regularization terms but has been proved to behave poorly in many situations. Qatten~\cite{yang2020qatten} employs a multi-head attention mechanism to compute the weights for the local action value functions and mix them to approximate the global Q-value. 
All of these methods aim to learn a value decomposition from the total reward signals to the individual value functions, which suffer from several problems: (i) The performance of the model highly relies on the decomposition function. (ii) The lacking of interpretability for the distributed credits. (iii) The high risk of the joint policy tends to fall into sub-optimal results~\cite{williams1991function,mnih2016asynchronous,ahmed2019understanding}. 

\subsection{Explicit Credit Assignment}
Explicit methods attribute the contributions for each agent that are at least provably locally optimal. The most representative method is COMA~\cite{foerster2018counterfactual}, which utilizes a counterfactual advantage baseline to guide the learning of local policies. However, it treats each agent as an independent unit and overlooks the complex correlations among agents. Thus, it becomes inefficient when encounters complex situations. 
SQDDPG~\cite{wang2020shapleyq} proposes a network to estimate the marginal contribution, which is further used to approximate Shapley Value. Then, Shapley Value is used to guide the learning of local agents. However, such estimation for marginal contribution doesn't make sense in many situations because the network over-relies on the assumption that the agents take actions sequentially.
QPD~\cite{yang2020q} designs a multi-channel mixer critic and leverage integrated gradients to distribute credits along paths, which achieves state-of-the-art results in many tasks. 
Intuitively, mining the relations between the agents is essential for the policy gradient in cooperative games. But the correlations are too complicated and are often underestimated by the models. To this end, we propose a Shapley Counterfactual Critic for credit assignment in MARL. Thanks to Shapley Value, the relations between the agents are considered sufficiently without prior knowledge, which further promotes the learning of local agents.
Different from SQDDPG~\cite{wang2020shapleyq}, we compute the marginal contributions according to a counterfactual method rather than building a network, which is more stable and efficient in complicated situations. 

\subsection{Shapley Value and Approximate SV}
Shapley Value~\cite{shapley1953value, bilbao2000shapley,meng2012core, sundararajan2020many} originates from cooperative game theory in the 1950s, which assigns a unique distribution of total benefits generated by the coalition of all players. Shapley Value satisfies the properties of \emph{efficiency}, \emph{symmetry}, \emph{nullity}, \emph{linearity} and \emph{coherency}. It is a unique and fairly way to quantify the importance of each player in the overall cooperation and widely used in economics.
However, the computational complexity of Shapley Value grows factorially with respect to the number of participating players~\cite{kumar2020problems}. Thus, in order to decrease the computation, several recent studies start to approximate the exact Shaply Value~\cite{fatima2008linear,chen2018shapley,ghorbani2019data,wang2021shapley} by sacrificing some properties. For example, Frye~\etal~\cite{frye2020asymmetric} and Tom~\etal~\cite{heskes2020causal} utilize casual knowledge to simplify its calculation, which breaks the axiom of \emph{symmetry}. L-Shapley and C-Shapley only consider the interactions among the local and connected player, which slightly break the properties of \emph{efficiency}. DASP~\cite{ancona2019explaining} and Neuron Shapley~\cite{ghorbani2020neuron} adapt sample methods to approximate Shapley Value, which also slightly breaks the properties of\emph{efficiency} and \emph{symmetry}.

\section{Preliminaries} \label{sec:preliminaries}
\subsection{Dec-POMDPs}
A fully cooperative multi-agent sequential decision-making task with $n$ agents $A = \{1, 2, ..., n\}$ can be modeled as a decentralised partially observable Markov decision process (Dec-POMDP)~\cite{oliehoek2016concise,bernstein2002complexity,busoniu2008comprehensive,gupta2017cooperative,palmer2018lenient}. 
Dec-POMDP is canonically formulated by the tuple:
\begin{equation*}
    G = ( S, U, P, r, Z, O, n, \gamma).
\end{equation*}
In the process, $s \in S $ represents the true state of the environment. At each time step, each agent $a \in A$ chooses an action $u_{a} \in U$ simultaneously to formulate a joint action   The action produces a state transition on the environment which is described by the Markov transition probability function $P(s'|s,u)$: $S \times U^{n} \rightarrow S$. All of the agents share a same global reward function $r(s,u)$: $S \times U^{n} \rightarrow \mathbb{R}$. 

In the setting of partial observability, the observations of each agent $z \in Z$ are generated by a observation function $O(s,a)$: $S \times A \rightarrow Z$. Each agent owns an action-observation history $\tau_{a} \in T$, where $T = (Z \times U)^{*}$ denotes the set of sequences of state-action pairs with arbitrary length. On this history, each agent conditions a stochastic policy $\pi^{a}(u_{a}|\tau_{a})$: $T \times U \rightarrow [0,1]$. The common goal of all agents is to maximize the expected discounted return $R_{t}={\sum}_{i=0}^{\infty}\gamma^{i}r_{t+i}$.

\subsection{Shapley Value}
Assume a coalition consists of $N$ players and they cooperate with each other to achieve a common goal. For a particular player $i$, let $S$ be a random set that contains player $i$ and $S \backslash \{i\}$ represents the set with the absence of $i$, then the marginal contribution of $i$ in $S$ is defined as:
\begin{equation}
\Delta v(i,S)=v(S)-v(S \backslash \{i\}),
\end{equation} 
where $v(\cdot)$ refers to the value function for estimating the cooperated contribution of a set of players.

Then the Shapley Value of player $i$ is computed as the weighted average of the marginal contributions in all of the subsets of $N$:
\begin{equation}
\phi_v(i)=\frac{1}{N}\textstyle{\sum}^N_{k=1} \frac{1}{\binom{N-1}{k-1}}\textstyle{\sum}_{S \in S_k(i)}{\Delta v(i,S)},
\end{equation}
where $S_k(i)$ denotes a set with size $k$ that contain the player $i$.
Shapley Value satisfy the following properties:

\begin{itemize}[leftmargin=*]
\item \textbf{Efficiency.} The credits generated by the big coalition $v(\{1,...,N\})-v(\emptyset)$ is equal to the sum of the Shapley Values of all of the participating players $\sum{^{N}_{i=1}}\phi_v(i)$.
\item \textbf{Symmetry.} If $\Delta v(i,S)=\Delta v(j,S)$ for all subsets $S$ then $\phi_v(i)=\phi_v(j)$.
\item \textbf{Nullity.} If $\Delta v(i,S)=0$ for all subsets $S$ then $\phi_v(i)=0$.
\item \textbf{Linearity.} Let $u$ and $w$ represent the associated gain functions, then $\phi_{v+w}(i)=\phi_v(i)+\phi_w(i)$ for every $i \in N$.
\item \textbf{Coherency.} When another value function $\Delta v'(i)$ is utilized to measure the marginal contribution of $i$, if $\Delta v(i,S)\geq \Delta{v'}(i,S)$ for all subsets $S$, then $\phi_v(i)\geq\phi'_v(i)$.

\end{itemize}
\subsection{Explicit Framework for MARL}
Explicit methods are interpretable for the allocated credits, which can reduce the suspicion of users to the rationality of the learned local agents.
For this reason, we extend the explicit methods~\cite{foerster2018counterfactual,wang2020shapleyq,yang2020q}, which first train the central critic according to the joint states and actions and then distribute the global reward signals according to the contributions of local agents to the critic. 

\begin{figure*}[ht]
    \centering
    \includegraphics[width=0.90\linewidth]{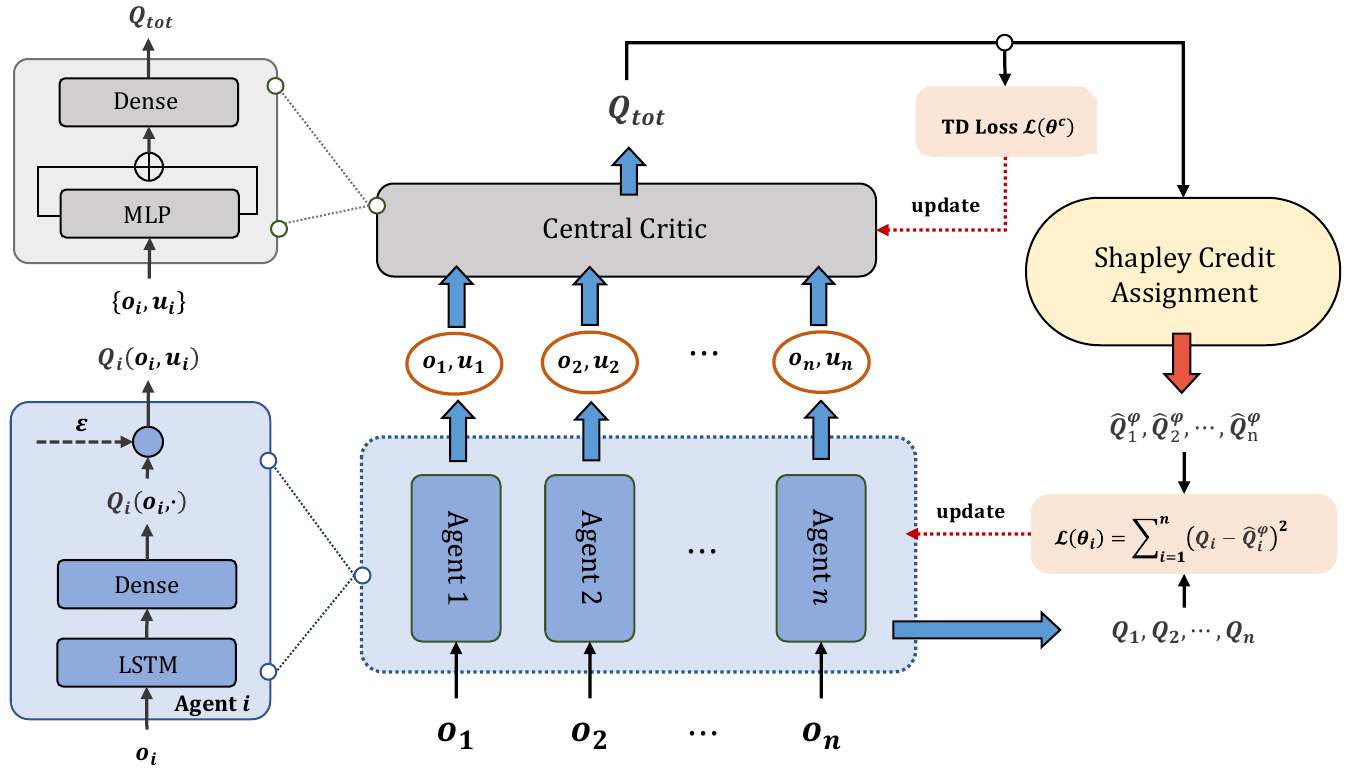}
    \caption{The framework of our method. We adopt a two-stage way that trains the central critic and the local policies separately. First, the central critic is updated with TD-loss. Then the credits of each agent are calculated by our proposed counterfactual method with approximate Shapley Value. Finally, the local policies are updated using the Shapley Counterfactual Credits.}
  \label{fig:framework}
\end{figure*}

Following QPD~\cite{yang2020q}, we model our critic network with three components as shown in Figure~\ref{fig:framework}, that is, the feature extraction module, the feature fusion module, and the Q-function estimation module. The first module consists of 2 dense layers with ReLU non-linearity, which is used to extract the features of a particular agent's observations and actions. Then
the features of all agents are concatenated thus merged into a global feature. Finally, the joint Q-value is computed according to the global feature.
As Yang~\etal~\cite{yang2020q} illustrated, different agents may own the same attributions, so can be categorized into different groups. For this reason, the agents within the same group are modeled using the same sub-network. Meanwhile, in order to simplify the network architecture and
accelerate the learning procedure, the agents of the same group share the same parameters.
We represent the central critic as:
\begin{equation}
    Q_{tot}=f(o_{1}, u_{1}, ..., o_{n}, u_{n}),
\end{equation}
where $o_{i}$ and $u_{i}$ denote the observation and the action of the $i$-th agent, respectively.

In our implementation, each local agent is realized with a Recurrent Deep Q-Network, which is composed of an Long Short-Term Memory (LSTM) layer and a Multi-Layer Perceptron (MLP). We represent the local agent as:
\begin{equation}
    Q_{i}=g(o_{i};h),
\end{equation}
where $h$ is the hidden state of LSTM.
For the exploration policy, $\epsilon$-greedy is adopted and the exploration rate of episode $eps$ is:
\begin{equation}
\epsilon(eps)=\max(\epsilon_{start}-eps \cdot \sigma, 0),
\end{equation}
where $\epsilon_{start}$ is the initial exploration rate and $\sigma$ represents the decreasing count of $\epsilon$ each episode.

\section{Shapley Counterfactual Credits for MARL} \label{sec:SCC}
The framework of our approach is illustrated in Figure~\ref{fig:framework}. First, the central critic takes the actions and observations of each agent as input and approximates the total Q value. Then the contributions of the individual agents are distributed
by the counterfactual method with Shapley Value. Finally, the local agents update their parameters according to the credits they earned.

In this section, we systematically describe our ``Shapley Counterfactual Credits'' for Multi-Agent Reinforcement Learning. First, we will introduce a counterfactual method with Shapley Value to address the problem of ``credit assignment'', which can fully mine the correlations among the local agents in \emph{Section~\ref{sec:4-1}}. To downgrade the computational complexity. we replace the truly Shapley Value with its approximation, and this will be discussed in \emph{Section~\ref{sec:4-2}}. The details of the proposed algorithm and the loss function will be introduced in \emph{Section~\ref{sec:4-3}}. 

\subsection{Counterfactual Method with Shapley Value for Credit Assignment} \label{sec:4-1}
The main challenge we need to address is how to measure the contributions of the agent. In other words, we need to quantify how the agents' actions influence the output of the central critic. COMA~\cite{foerster2018counterfactual} proposed a special critic and utilize a counterfactual baseline, which estimated the advantages of action value over expected value as this influence but shows poor performance on many tasks. Wolpert~\etal~\cite{wolpert2002optimal} computed the influence by using \emph{difference rewards} which compares the global reward to the reward received when the action of an agent is replaced with a default action. Inspired by these ideas, we also proposed a counterfactual method in our central critic to measure the effect of the actions taken by the agents. 

We consider the contribution of an action taken by an agent is equal to ``how the output will change when this action is absent?" We formulate the contributions of the action performed by the $i$-th agent to the central critic as:
\begin{equation}
    v_{i}=f(o_{1}, u_{1}, ...,o_{i}, u_{i}, ..., o_{n}, u_{n}) - f(o_{1}, u_{1}, ...,o_{i}, \Tilde{u}, ..., o_{n}, u_{n}),
    \label{cf}
\end{equation}
where $\Tilde{u}$ denotes a baseline that means the action is replaced by a default one.

However, such estimation for the contributions is insufficient since the agents are cooperating with each other and cannot be treated as independent units. We then desire to quantify the credits made by an agent precisely from the intricate relationship among agents, but the environment is complex, and there is no prior knowledge to indicate how they cooperated with each other. To this end, we propose to utilize Shapley Value for credit assignment, and this will be introduced in the next subsection.

As we mentioned before, Shapley Value distributes the credits fairly by considering the contributions of the participating players and satisfies many good properties such as \emph{efficiency}, \emph{additivity}, and \emph{coherency}. Thus, we utilize this tool to extend the counterfactual method.

For convenience, we shorthand Equation~\eqref{cf} and change agent $i$ to a set $S$ of agents:
\begin{equation}
    v_{S~in~A}= f(H_{A})-f(H_{A\backslash S}),
\end{equation}
where $A$ denotes all of the agents, $H_{A}$ represents the actions and observations of $A$, and $H_{A\backslash S}$ denotes that the actions of all agents in $S$ are replaced with default actions.

To compute the Shapley Value of the $i$-th agent in the big coalition, we need to compute its marginal contributions when this agent play roles in all of the subset of the big coalition $A$. We define the marginal contribution of the $i$-th agent in the subset $S$ of $A$ as:
\begin{equation}
    \Delta v(i,S)= v_{S~in~A} - v_{S\backslash i~in~A},
\end{equation}
where $S\backslash i$ denotes $S$ with the removal of the $i$-th agent.

After getting the marginal contribution, we compute the Shapley Counterfactual Credits $Q^{\varphi}_{i}$ as:
\begin{equation}
Q^{\varphi}_{i}=\frac{1}{N}\textstyle{\sum}^N_{j=1} \frac{1}{\binom{N-1}{j-1}}\textstyle{\sum}_{S \in S_j(i)}{\Delta v(i,S)},
\end{equation}
where $S_j(i)$ denotes the set of agents with size $j$ that contains the $i$-th agent.

\subsection{Approximation of Shapley Value} \label{sec:4-2}
However, the main drawback of Shapley Value is that the computational complexity grows factorially as the number of the agents increases~\cite{kumar2020problems}. So recent studies usually use an approximation of Shapley Value as a substitution~\cite{chen2018shapley,ghorbani2019data,wang2021shapley}. Since the number of the agents may bring an unacceptable computational cost, for alleviating the computational burden, the approximation of Shapley Value is necessary. Thus, we adopt the Monte Carlo sampling method to get the approximated Shapley Value:
\begin{equation}
\hat{Q}^{\varphi}_{i}=\frac{1}{M}\textstyle{\sum}^{M}_{j=1}{\Delta v(i,S_{MC_j}(i))},
\label{ASC}
\end{equation}
where $M$ represents the times of Monte Carlo sampling, $S_{MC_j}(i)$ represents a subset of $A$ sampled in $j$-th time that contains the $i$-th agent.

According to this approximation, we downgrade the computational complexity of the truly Shapley Value of an agent from $O(N!)$ to $O({M})$, where $N$ is the number of agents that may be very large in some situations, and $M$ is a hyperparameter which represents the times of Monte Carlo sampling and can be a small positive integer.
To be noticed that, such an approximation of Shapley Value might slightly break some of its properties such as \emph{efficiency} and \emph{Symmetry}. Recent literature sacrificed its properties in varying degrees but got an acceptable computational costs~\cite{chen2018shapley,ghorbani2019data,wang2021shapley}. We deem that such an approximation is necessary and will not bring too much impact to the model's performance.

\subsection{Loss Function and Training Algorithm} \label{sec:4-3}
We show the details of our algorithm in Algorithm~\ref{alg}.
Our whole framework is updated in two stages. 
First, the local agents interact with the environment and take actions according to their observations and history. Then, these actions and observations act as the input of the central critic to estimate the joint Q-function. Afterward, in the first stage, we update the central critic by minimizing the TD-loss $\mathcal{L}(\theta^c)$:
\begin{equation}
  \begin{split}
    \mathcal{L}(\theta^{c})=(Q_{tot}-y)^{2}, \\
    y=r+\gamma(\tilde{Q}_{tot}), \\
    \end{split}
    \label{TD}
\end{equation}
where $\theta^{c}$ is the parameters of the central critic, $Q_{tot}$ is the output of the central critic, and $\tilde{Q}_{tot}$ represents the output of target network of the central critic.

In the second stage, we first get the Shapley Counterfactual Credits $Q^{\varphi}_{i}$ of each agent according to Equation~\eqref{ASC}. Then each agent is trained by minimizing the loss:
\begin{equation}
\mathcal{L}(\theta^{i})=(Q_{i}-\hat{Q}^{\varphi}_{i})^{2},
\label{LL}
\end{equation}
\noindent where $\theta^{i}$ denotes the parameters of the $i$-th local agent, and $Q_{i}$ is the output of the $i$-th agent.
\begin{algorithm}

\caption{Shapley Counterfactual Credits Algorithm for MARL}
\hspace*{0.02in} {\bf Initialize:} 
Central critic network $\theta^{c}$, target central critic network $\tilde{\theta}^{c}$, local agents' networks $\theta^{\pi}=(\theta^{1},...,\theta^{n})$
\begin{algorithmic}[1]
\For{each training episode $eps$} 
\State $s_{0}$ = initial state, $t$ = 0, $h^{i}_{0}$  = 0 for each agent $i$
    \While{{$s \neq$ terminal} \textbf{and} {$t<T$} }
    \State $t=t+1$
    \For{each agent $i$}
    \State $Q_{i}(o^{i}_{t},·), h^{i}_{t}=Agent_{i}(o^{i}_{t};h^{i}_{t-1})$
    \State Sample $u^{i}_{t}$ from $\pi(Q_{i}(o^{i}_{t},·),\epsilon(eps))$
    \EndFor
    \State Execute the joint action $(u^{1}_{t},u^{2}_{t},...,u^{n}_{t})$
    \State Get reward $r_{t+1}$ and next state ${s_{t+1}}$
    \EndWhile
    \State Add episode to replay buffer
    \State Collate episodes in buffer into a single batch
    \For{$b$ in batch}
        \For{$t=1$ to $T$}
        \State Compute the targets $y^{t}$ using central target network
        \EndFor
    \EndFor
    \State Update central critic network $\theta^{c}$ with \eqref{TD}
    \State Every $C$ episodes reset $\tilde{\theta}^{c}=\theta^{c}$
    \For{$b$ in batch}
        \For{$t=1$ to $T$}
        \State Compute credits for each agent via \eqref{ASC}
        \EndFor
    \EndFor
    \State Update the local agents $\theta^{\pi}$ with \eqref{LL}
\EndFor 
\end{algorithmic}
\label{alg}
\end{algorithm}

\begin{figure*}[h]
    \begin{subfigure}{\textwidth}
        \includegraphics[width=\linewidth]{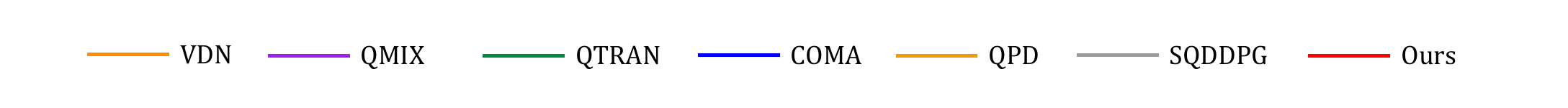}
    \end{subfigure}
    \begin{subfigure}{0.33\textwidth}
    \includegraphics[width=\linewidth]{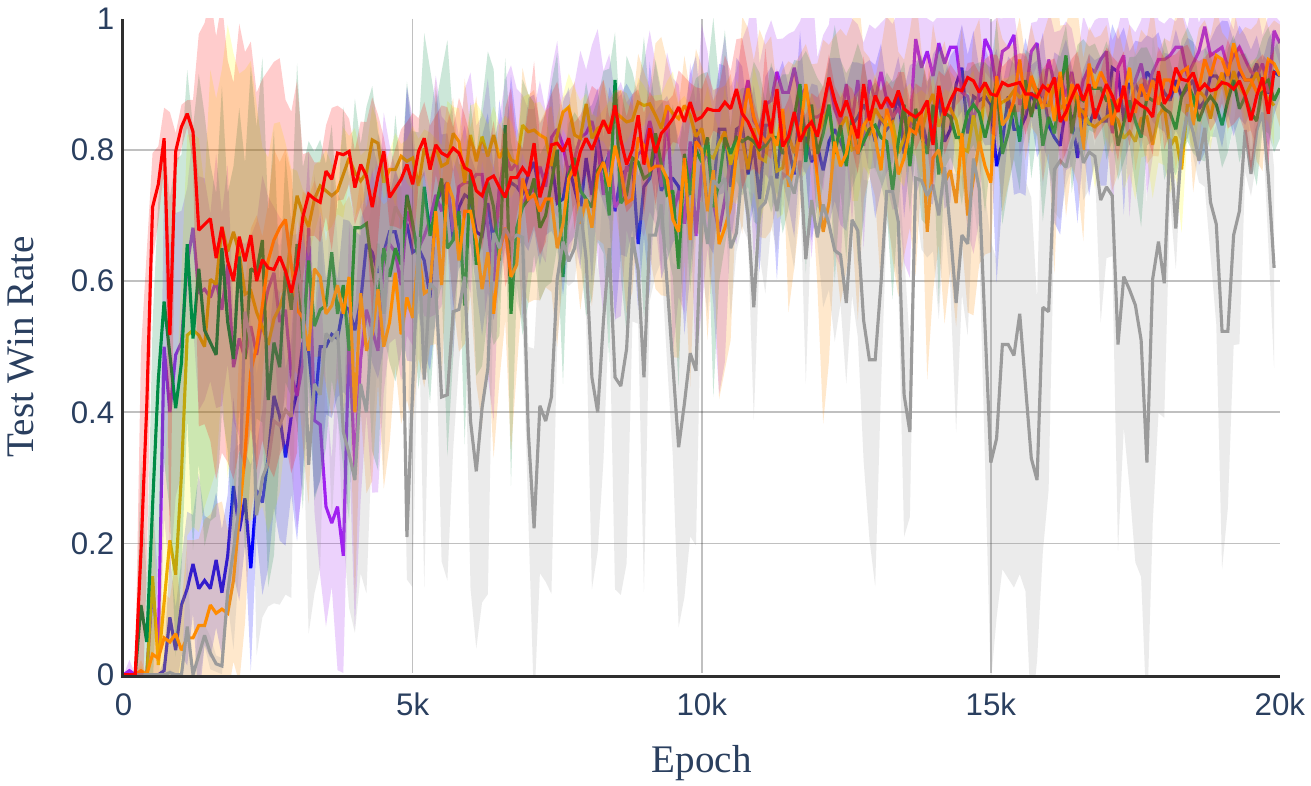}
    \caption{3m}    
    \end{subfigure}
    \begin{subfigure}{0.33\textwidth}
        \includegraphics[width=\linewidth]{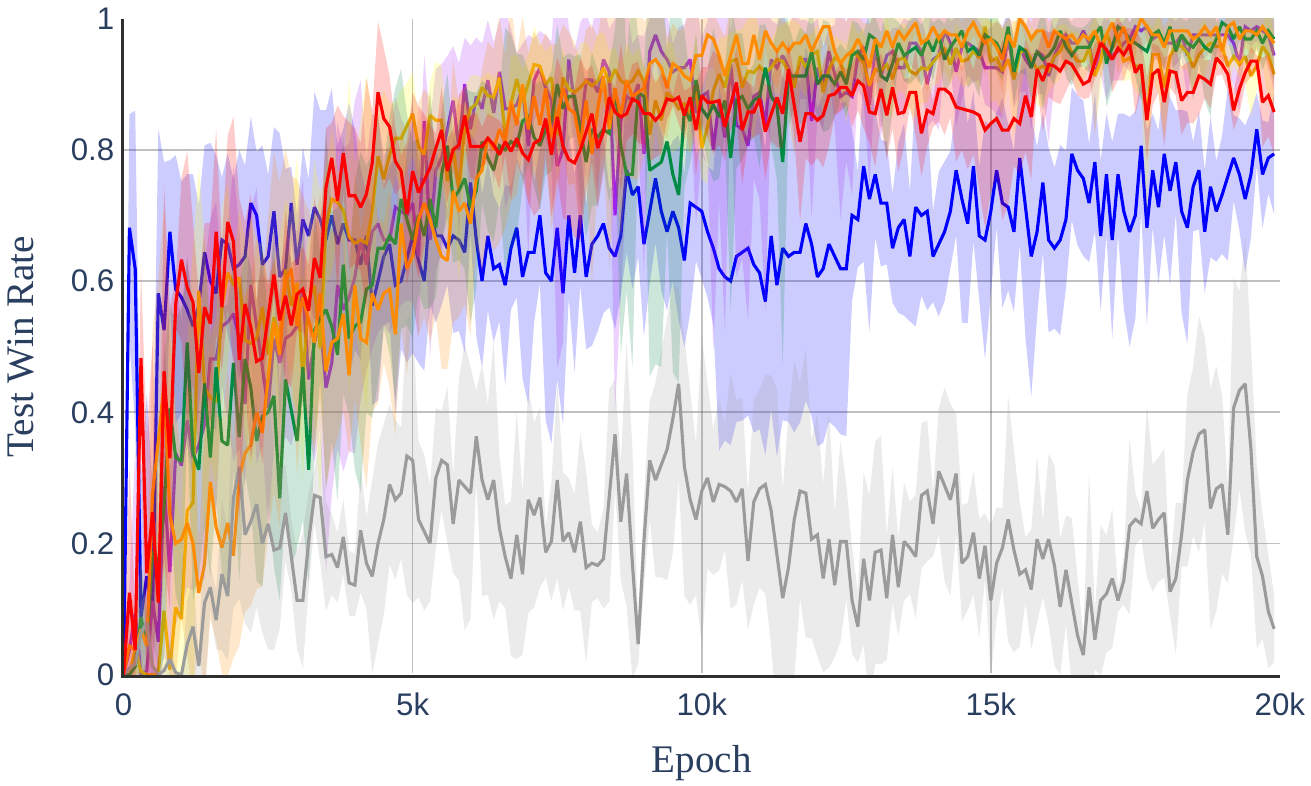}
        \caption{8m}
    \end{subfigure}
    \begin{subfigure}{0.33\textwidth}
        \includegraphics[width=\linewidth]{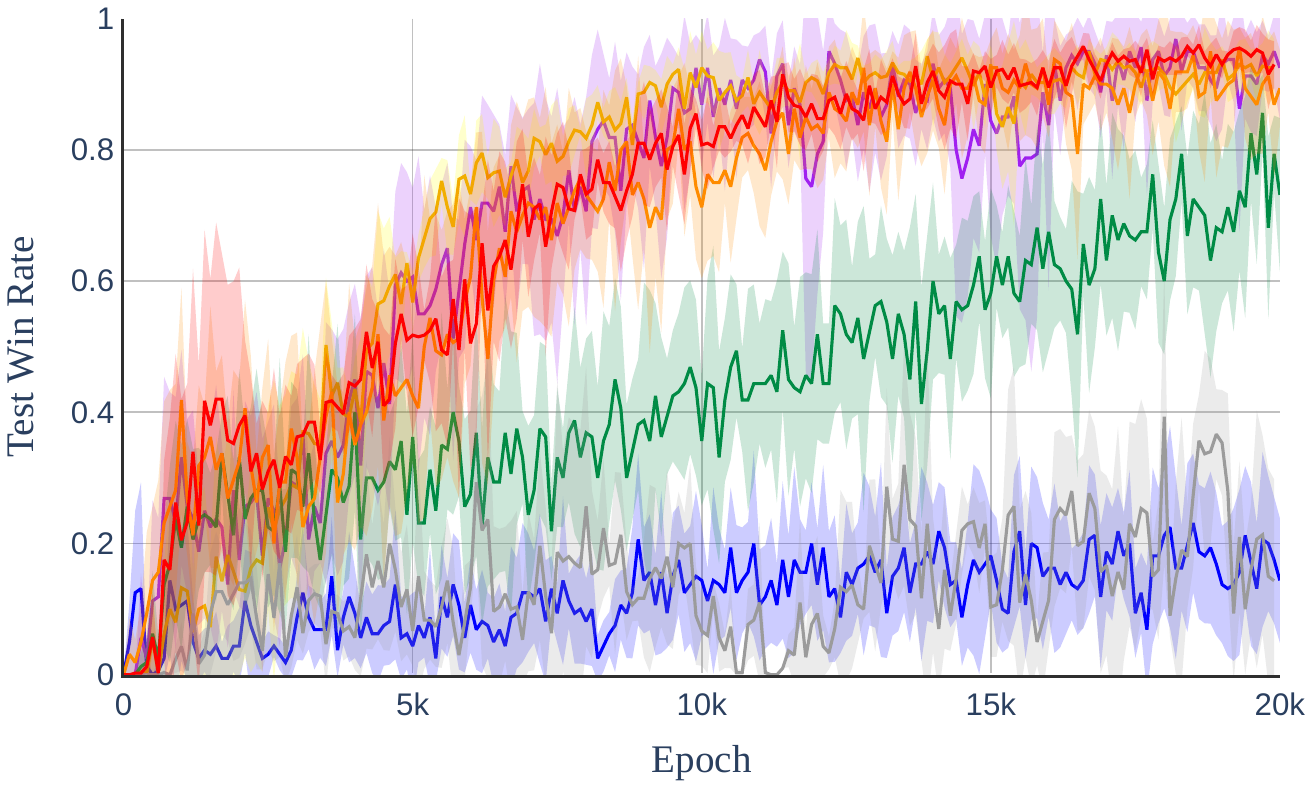}
        \caption{2s3z}
    \end{subfigure}
   \begin{subfigure}{0.33\textwidth}
    \includegraphics[width=\linewidth]{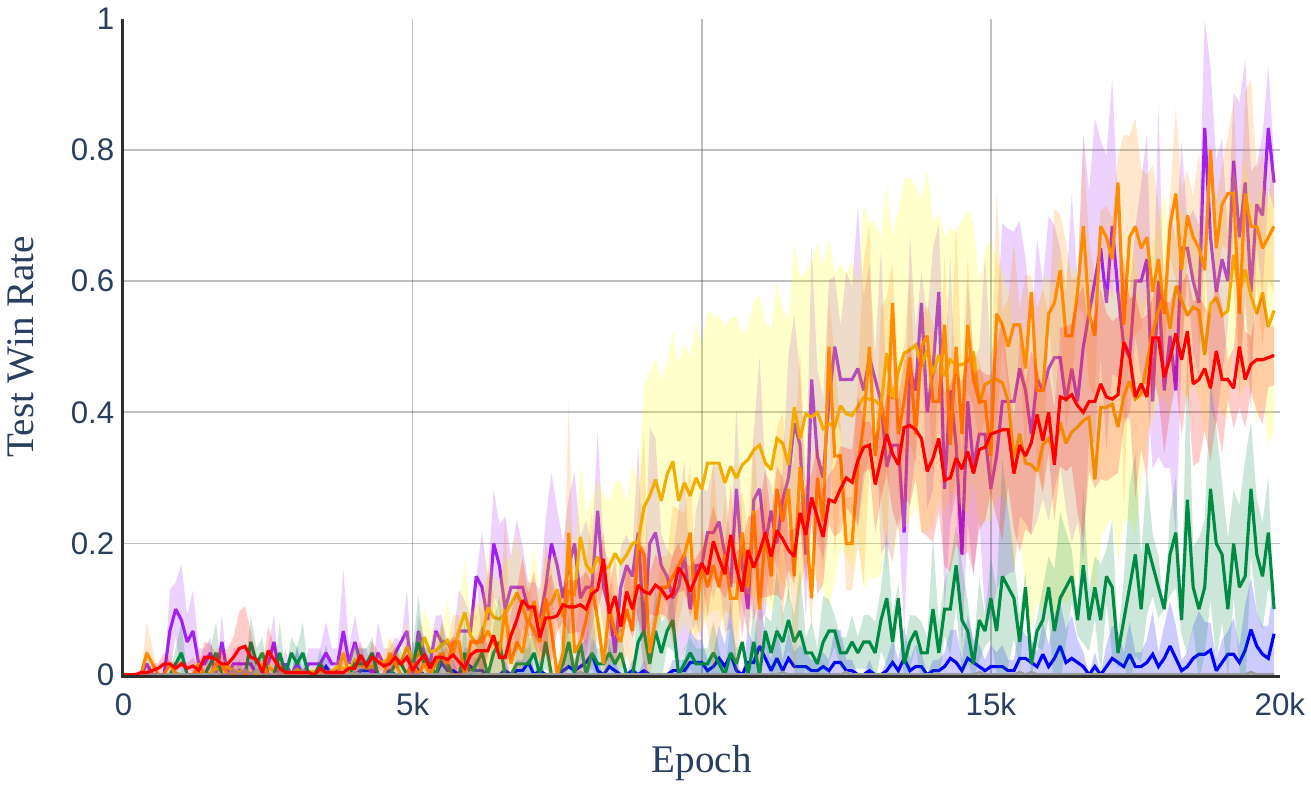}
    \caption{1c3s5z}    
    \end{subfigure}
    \begin{subfigure}{0.33\textwidth}
        \includegraphics[width=\linewidth]{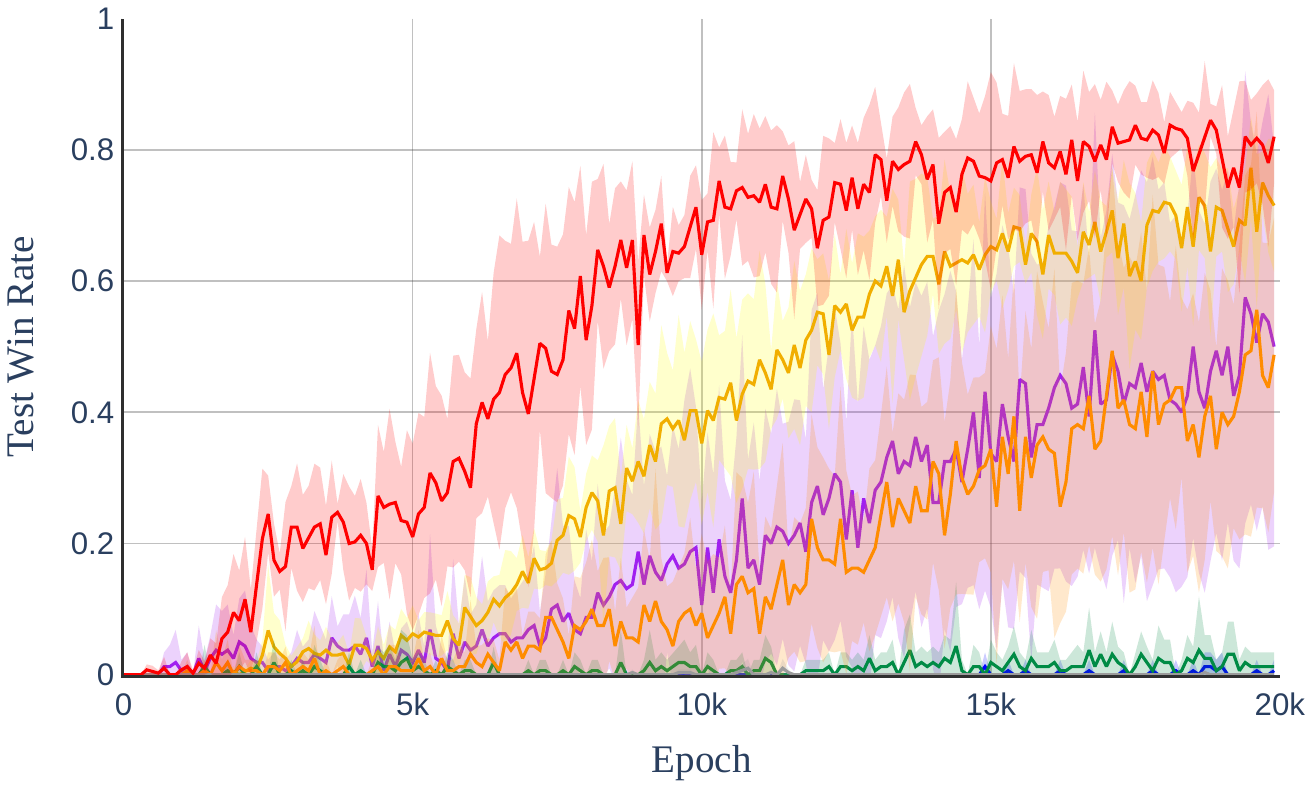}
        \caption{3s5z}
    \end{subfigure}
    \begin{subfigure}{0.33\textwidth}
        \includegraphics[width=\linewidth]{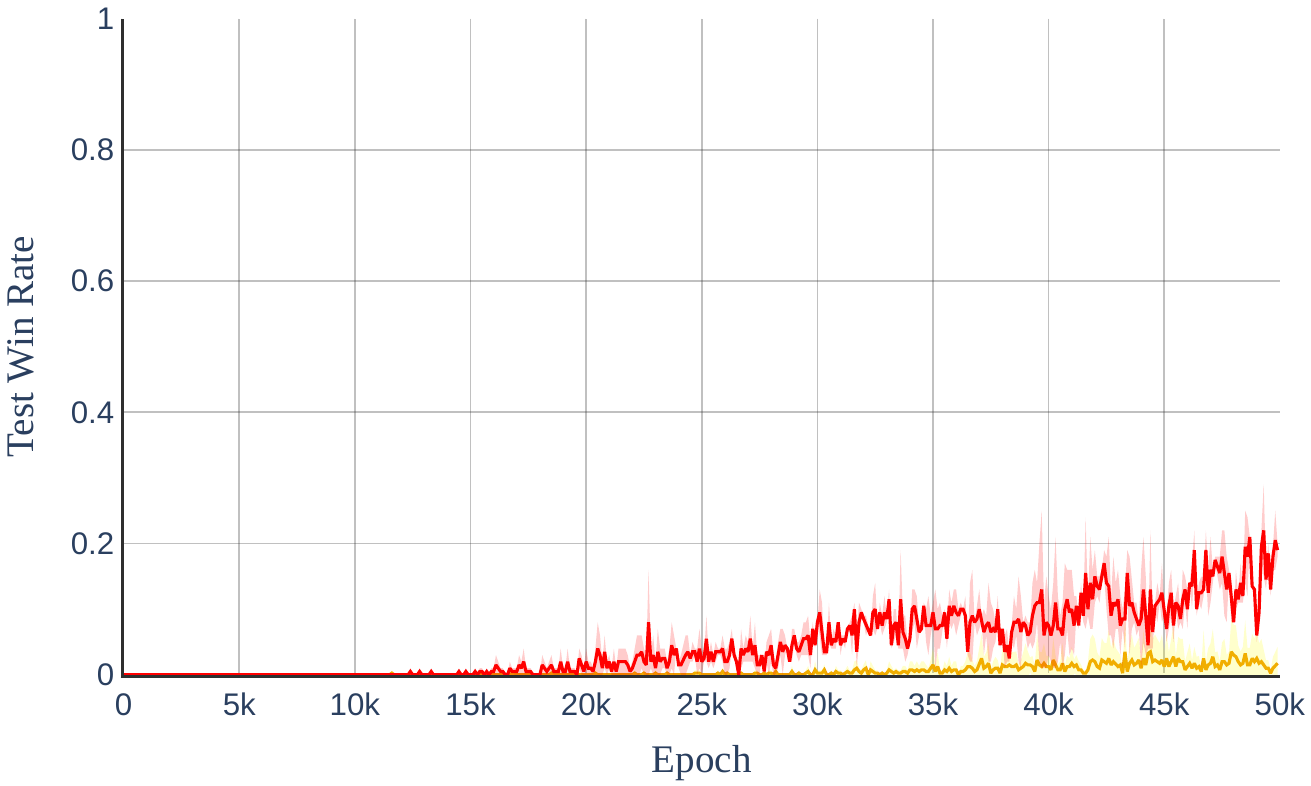}
        \caption{3s5z\_vs\_3s6z}
    \end{subfigure}
    \caption{The mean win rates of our method compared with others in different map scenarios of StarCraft II. The shaded areas represent the standard deviation.}
    \label{fig:results}
\end{figure*}

\begin{table*}[t]
\caption{Median and mean win rate of our method compared with other methods. $\tilde{m}$ represents the median of the test win rates and $avg$ represents mean test win rates.}
\fontsize{9}{12}\selectfont
\begin{tabular}{|l|cc|cc|cc|cc|cc|cc|cc|}
\hline
\multirow{3}{*}{Map}            & \multicolumn{14}{c|}{Methods } \\  \cline{2-15}
& \multicolumn{2}{c|}{VDN}    & \multicolumn{2}{c|}{QMIX}   & \multicolumn{2}{c|}{QTRAN}  & \multicolumn{2}{c|}{COMA}  & \multicolumn{2}{c|}{QPD} & \multicolumn{2}{c|}{SQDDPG} & \multicolumn{2}{c|}{OURS}  \\ 
& $\tilde{m}$ &$avg$   &$\tilde{m}$ &$avg$   &$\tilde{m}$ &$avg$  &$\tilde{m}$ &$avg$  &$\tilde{m}$ & $avg$  &$\tilde{m}$ &$avg$ &$\tilde{m}$ &$avg$ \\ 
\hline
3m  &\textbf{100} &\textbf{100} &\textbf{100} &\textbf{100}  &\textbf{100}  &\textbf{100} &95 &96 &99 &99 &64 &65 &99 &99  \\
8m &\textbf{100} &\textbf{100} &\textbf{100} &\textbf{100} &\textbf{100} &\textbf{100} &\textbf{100} &\textbf{100} &95 &95 &92 &90 &98 &97 \\
2s3z &\textbf{100} &\textbf{100} &\textbf{100} &\textbf{100} &92 &91 &45 &45  &99  &98  &60 &55 &\textbf{100} &\textbf{100} \\
1c3s5z & 88 & 85 &\textbf{95} &\textbf{90} & 40 & 41 & 15 & 15 &77 &72 &2 &2 &61 &60 \\
3s5z & 80 & 69  &80 & 67 & 12 & 13 &5 &3 &79 & 80 &1 &1 &\textbf{92} &\textbf{90} \\
3s5z\_vs\_3s6z &0 &0 &0 &0 &0 &0 &0 &0 &3 &5 &0 &0 &\textbf{20} &\textbf{20} \\
\hline
\end{tabular}
\label{tab:final_results}
\end{table*}

\begin{figure*}[h]
    \begin{subfigure}{\textwidth}
        \includegraphics[width=\linewidth]{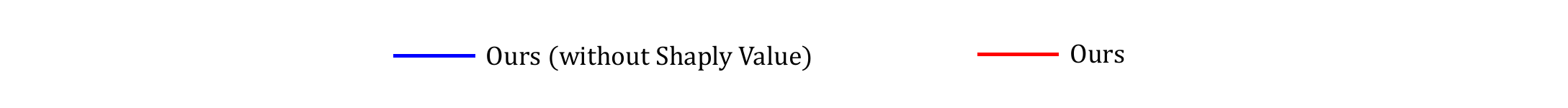}
    \end{subfigure}
    \begin{subfigure}{0.33\textwidth}
    \includegraphics[width=\linewidth]{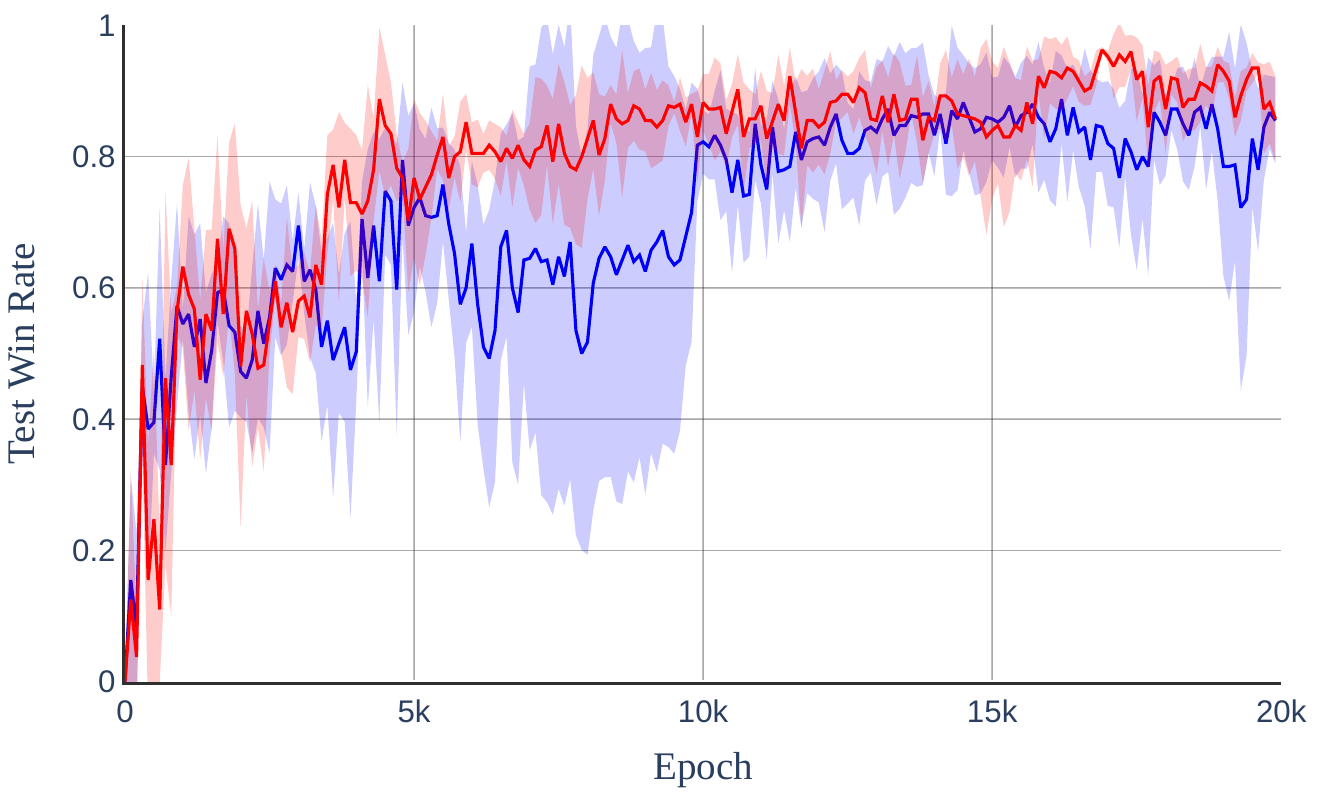}
    \caption{3m}    
    \end{subfigure}
    \begin{subfigure}{0.33\textwidth}
        \includegraphics[width=\linewidth]{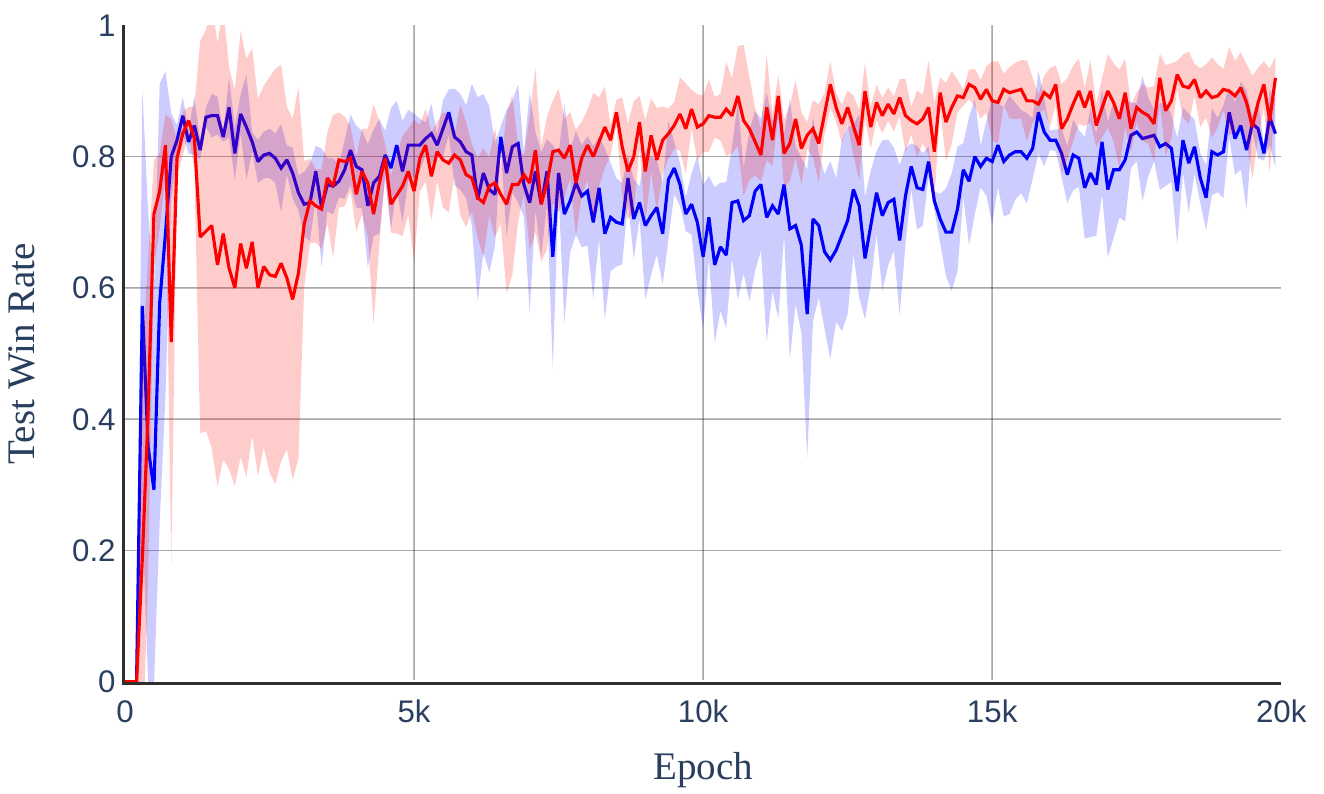}
        \caption{8m}
    \end{subfigure}
    \begin{subfigure}{0.33\textwidth}
        \includegraphics[width=\linewidth]{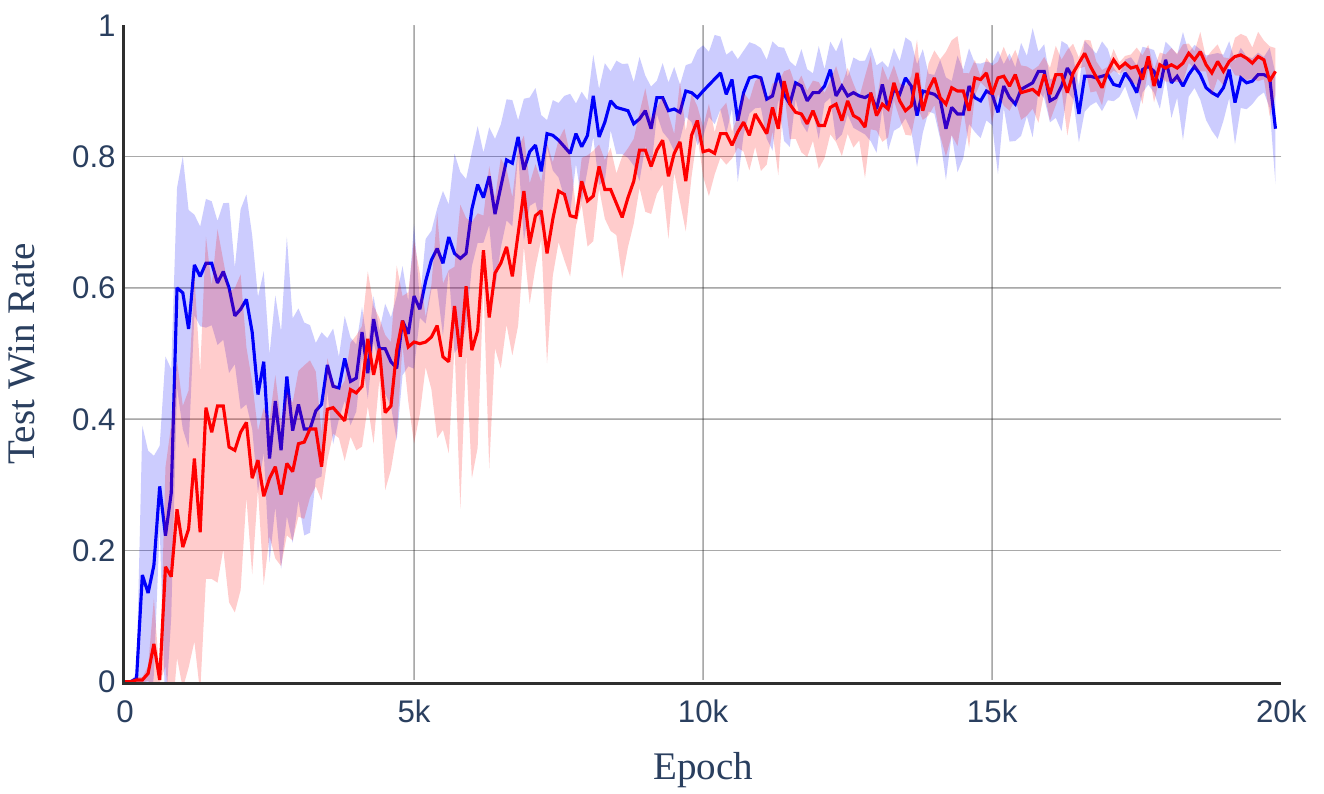}
        \caption{2s3z}
    \end{subfigure}
    \caption{Ablation study of Counterfactual Shapley Credits.}
\label{fig:ablation}
\end{figure*}

\section{Experiments} \label{sec:experiments}
We focus on addressing the problem of credit assignment in MARL with cooperative settings explicitly. We compare our proposed method with several baselines, including VDN~\cite{sunehag2018value}, QMIX~\cite{rashid2018qmix}, COMA~\cite{foerster2018counterfactual}, QTRAN~\cite{son2019qtran}, QPD~\cite{yang2020q}, and SQDDPG~\cite{wang2020shapleyq}. The training configurations, experiment results, as well as the analysis will be described in detail in this section.

\subsection{Experiment Settings}
\paragraph{Environment}
We perform extensive experiments on the StarCraft II (a real-time strategy game) micromanagement challenge, in which each army is controlled by an agent and act based on its local
observations and the opponent's army are controlled by the hand-coded built-in StarCraft II AI.
Each unit in StarCraft contains a rich set of complex micro-actions, which allow the learning of complex interactions between the agents that cooperate with each other.
The overall goal is to maximize the accumulated rewards for each battle scenario. The environment produces rewards based on the hit-point damage dealt and enemy units killed. Besides, another bonus is given when the battle wins. 
At each time step, each agent can only receive the local observations within its field of view. Meanwhile, an agent can only observe the other agents alive and located in its sight range. 
Besides, all agents can only attack the enemies within their shooting range, which is set to 6. 
The global state consists of the joint observations without the restriction of the sight range, which will be used in the central critic during the training procedure.
All features are normalized by their maximum values before sent to the neural network.
StarCraft Multi-Agent Challenge (SMAC) environment~\cite{samvelyan2019starcraft} is used as testbed, and we set the difficulty of the game AI as ``very difficult'' level.

\paragraph{Configurations}
The central critic of our method is the same as QPD~\cite{yang2020q}, which consists of the feature extraction layers, the feature fusion operation, and the Q-function estimation layers. First, the agents are grouped according to their attributions, and 2 dense layers are used to extract the features of their observations and actions. Each dense layer consists of 64 neurons for each channel. For accelerating the learning procedure, we adopt parameter sharing technique~\cite{yang2018mean,iqbal2019actor} where the agents within the same group share the parameters of the feature extraction layers. Then, we concatenated the features of all agents to fuse them into a global feature. Finally, for the final Q-function estimation, we adapt another dense layer with one output neuron. In the procedure of computing Shapley Value, we adapt the Monte Carlo sampling method to sample 5 subsets for each agent at each time step. We set the counterfactual baseline $\Tilde{u}$ in the central critic as zero vector for convenience. 
We model the local agents with an LSTM layer and 2 fully connected layers. The dimensional of hidden state in LSTM is set as 64, the units of the two fully connected layers are set as 64 and $|U|$ separately, where $|U|$ is the size of action space.
We set the discount rate $\gamma$ for TD-loss as 0.99.  
The replay buffer stores the most recent 1000 trajectories.
During training, we update the central critic with Adam and local agent networks with RMSProp. We copy the parameters of the central critic to its target network every 200 training episodes. 
The full hyperparameters of our Shapley Counterfactual Credits are shown in Table~\ref{tab:hyper}. 
The map \emph{3s5z\_vs\_3s6z} is much harder than the other maps, and the allied forces have one unit less than the enemy. During training, the win rates remain 0 even when the returns are relatively high.
For this reason, we set the number of the training episodes of map \emph{3s5z\_vs\_3s6z} to 50000, while the others are set to 20000. 

\begin{table}[t]
\caption{Hyperparameters of Shapley Counterfactual Credit Algorithm}
\fontsize{9}{12}\selectfont
\begin{tabular}{|l|l|}
\hline
Settings & Value \\ \hline
Batch size & 32 \\
Replay buffer size & 1000 \\
Training episodes  & 20000 \\
Exploration episodes  & 1000 \\
Start exploration rate & 1 \\
End exploration rate & 0 \\
TD-loss discount & 0.9 \\
Target central critic update interval & 200 episodes \\
Evaluation interval & 100  episodes \\
Evaluation battle number & 100   \\
Agent optimizer & RMSProp \\ 
Central Critic optimizer & Adam \\ 
Agent learning rate & 0.0005 \\ 
Central critic learning rate  & 0.001 \\ 
Dense units & 64\\
LSTM hidden units & 64\\
Baseline for Shapley Value & 0 vector \\
Times for Monte Carlo Sampling & 5\\ \hline
\end{tabular}
\label{tab:hyper}
\end{table}

\subsection{Results and Analysis}
To demonstrate the efficiency of our proposed method, we perform experiments on 6 maps of StarCraft II (\emph{3m}, \emph{8m}, \emph{2s3z}, \emph{1c3s5z}, \emph{3s5z}, \emph{3s5z\_vs\_3s6z}), including both homogeneous and heterogeneous scenarios. Figure~\ref{fig:results} depicts the curve of mean win rates of our method compared to the baselines. The final results of our method are depicted in Table~\ref{tab:final_results}, where $\tilde{m}$ represents the median of the test win rates and $avg$ represents mean test win rates.

All of the methods show high performance on three simple scenarios (\emph{3m}, \emph{8m}, \emph{2s3z}), and our Shapley Counterfactual Credits algorithm is competitive with the state-of-the-art algorithm, and achieves nearly 100\% mean win rates. Both sides have 3 \emph{Marines} in map \emph{3m}, and 8 \emph{Marines} in map \emph{8m}. As the arms of both sides are single and the numbers are equal, each agent only needs to focus on beating enemies and avoid taking redundant actions.
Concretely, from the replay, in map \emph{3m} and \emph{8m}, units learned to stand in a line or semicircle in order to set fire to the incoming enemies.
Such a pattern is easy for models to learn, and agents hardly need to consider how to cooperate with its friendly forces. 
In map \emph{2s3z}, both sides have 2 \emph{Stalkers} and 3 \emph{Zealots}. Since that \emph{Zealots} counter \emph{Stalkers}, the \emph{Stalkers} need to hide behind the own side \emph{Zealots}. Such a small number of units does not bring too much challenge for the learning of the model.

Our algorithm falls behind the other methods in map \emph{1c3s5z}, where both sides have 3 \emph{Stalkers}, 5 \emph{Zealots} and an \emph{Colossus}. Since the \emph{Colossus} is more threatening, and becomes the priority target, which reduces the difficulty of the game. Here, we divide the learned ability of an agent into the personal ability and the cooperative ability. For example, ``kite the enemy'' as well as ``attack high-threat targets'' belongs to the former, and ``move to protect the allies'' belongs to the latter. In this map, all of the agents need to learn the pattern to attack the enemy's \emph{Colossus} first, which makes other actions less important. Since Shapley Value focuses more on mining the correlation between agents, our method does not perform very well in this scenario.

Our algorithm shows obvious advantages in two maps \emph{3s5z} and \emph{3s5z\_vs\_3s6z} which are much more difficult than others. In map \emph{3s5z}, both sides have 3 \emph{Stalkers} and 5 \emph{Zealots}, and we got the mean win rates of 90\%. In this scenario, not only the agents of \emph{Stalkers} need to stand behind the allied \emph{Zealots}, but learn to attack the enemy \emph{Stalkers} with high priority. Meanwhile, the allied \emph{Zealots} need to protect allied \emph{Stalkers} as well as attack the nearest enemy \emph{Stalkers}. 
In this complex situation, cooperation among agents is more important than before.
Our counterfactual method with Shapley Value fully considers the correlation and interactions between units and distributes a moderate credit for the actions taken by each agent, thus outperforms the baselines significantly. For instance, a ``movement'' of a \emph{Zealots} may affect other friendly forces in varying degrees; we measure its contribution by considering how the results will change when different kinds of correlations are absent. 
Especially, in map \emph{3s5z\_vs\_3s6z}, where ally has 3 \emph{Stalkers} and 5 \emph{Zealots} while the enemy has 6 \emph{Zealots}, all of the current method except QPD got the mean win rates of zero. 
The reason for the poor performance of these methods is that cooperative behavior such as ``block'' rather than ``kite'' play more important roles in such settings. The \emph{Zealots} need to attract firepower in order to protect the allied \emph{Stalkers}, which is the only way to get the final victory. In this scenario, Shapley value fully demonstrates its superiority.
Our method achieves the mean win rates of 20\%, and reach the state-of-the-art. 

In conclusion, our proposed Shapley Counterfactual Credits algorithm shows its strength and beats all of the other methods in complicated scenarios where cooperation among agents plays an essential role. Our proposed algorithm also exhibits the competitive results with the state-of-the-art algorithm in the scenarios that need to pay more attention to personal ability.

\subsection{Ablation Study}
To demonstrate the advantage of Shapley Value~\cite{shapley1953value} to the counterfactual method, we perform ablation study on three maps (\emph{3m}, \emph{8m}, \emph{2s3z}). The difficulty of these three maps increases sequentially. The results are shown in Figure~\ref{fig:ablation}. The blue curves represent that the credits are allocated by the counterfactual method without Shapley Value. The red curves represent that the credits are distributed by Shapley Counterfactual Credits. For the balance between the performance and computational costs, we set the times of the Monte Carlo sampling for approximating Shapley Value as 5, and the analysis is shown in the next subsection.

In map \emph{3m} and \emph{8m}, the units need to learn the strategies that stand in a suitable position to fire the same enemy unit together. Thus, the ability to cooperate is relatively important, and the use of Shapley Value brings an improvement of the performance. While in \emph{2s3z}, the \emph{Stalkers} need to ``kite the \emph{Zealots}'' and the number of the units is small, which means personal ability is more important. So our method loses advantage in this scenario. 
It is worth mentioning that the use of Shapley Value makes learning more stable and reduces the standard deviation (the shaded part in the figure) of the win rates significantly. That because Shapley Value considers a variety of combinations among agents and measure the contribution of an agent via the weighted average of the counterfactual results of these combinations.

\subsection{The Choice of Sample Times for Shapley Approximation}
\begin{figure}[t]
    \begin{subfigure}{1\linewidth}
    \centering
        \includegraphics[width=0.85\linewidth]{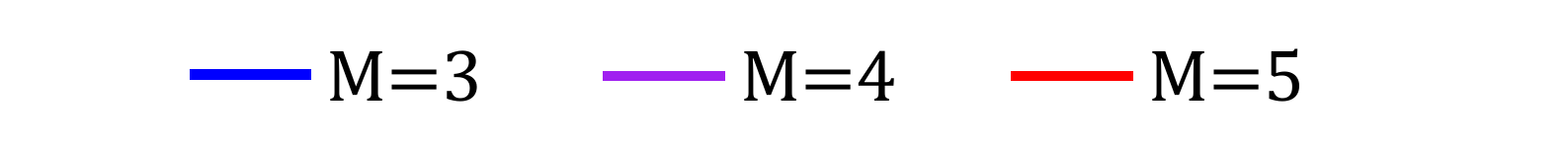}
    \end{subfigure}
    \begin{subfigure}{1\linewidth}\centering
        \includegraphics[width=0.95\linewidth]{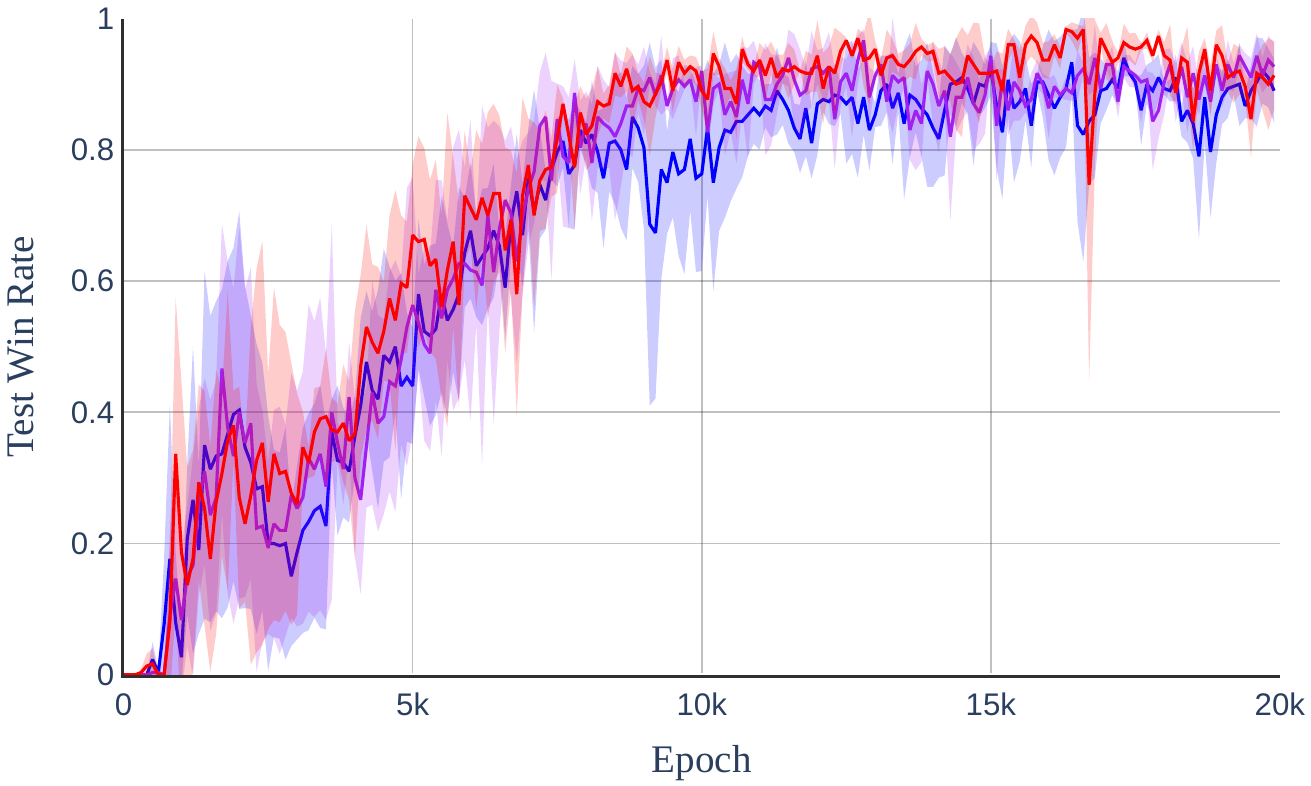}
    \end{subfigure}
    \caption{The mean win rates of the approximated Shapley Counterfactual Credits with different sample times in map \emph{2s3z}. }
\label{fig:|M|}
\end{figure}

We approximated the truly Shapley Value via Monte Carlo sampling.
Concretely, at each time step, we sample $M$ subsets randomly for each agent $a$, and average the marginal contributions of $a$ in these subsets to represent its approximated Shapley Value.
However, a large $M$ will still bring pressure to the computation costs, and small $M$ will lead to an inaccurate approximation. We performed extensive experiments to find a moderate hyperparameter, and the results are depicted in Figure~\ref{fig:|M|}. We conclude that 4 times sampling is sufficient to reach an ideal result. But to make the performance more stable, we set $M$ to 5 for in our experiments.

\section{Conclusion and Future Work} \label{sec:conclusion}
In this paper, we investigate the problem of credit assignment in Multi-Agent Reinforcement Learning. 
We extend the methods of explicit credit assignment and leverage a counterfactual method to measure the contributions of local agents to the central critic. 
To fully describe the relationships among the cooperative agents, Shapley Value is utilized with a sample-based method, with a Monte-Carlo sampling variant to decrease its computational complexity from factorial to polynomial. 
Experiments on the StarCraft II micromanagement tasks show the superiority of our method as we reach the state-of-the-art on various scenarios.

For future work, it could be interesting to investigate the causal knowledge among the cooperative agents. With this inferred knowledge, Shapley Value can be approximated in a more accurate way and the credit assignment can be more precise. Our method can also be extended to the scenarios with competitive settings, where variants of Shapley Value are proved to be effective.

\begin{acks}
This work was supported by the National Key Research and Development Project of China (No.2018AAA0101900), the National Natural Science Foundation of China (No. 61625107, U19B2043, 61976185, No. 62006207), Zhejiang Natural Science Foundation (LR19F020002), Key R \& D Projects of the Ministry of Science and Technology (No. 2020YFC0832500), Zhejiang Innovation Foundation(2019R52002), the Fundamental Research Funds for the Central Universities and Zhejiang Province Natural Science Foundation (No. LQ21F020020), Baoxiang Wang is partially supported by AC01202101031 and AC01202108001 from AIRS.
\end{acks}
\bibliographystyle{ACM-Reference-Format}
\bibliography{reference}


\begin{thebibliography}{45}


\ifx \showCODEN    \undefined \def \showCODEN     #1{\unskip}     \fi
\ifx \showDOI      \undefined \def \showDOI       #1{#1}\fi
\ifx \showISBNx    \undefined \def \showISBNx     #1{\unskip}     \fi
\ifx \showISBNxiii \undefined \def \showISBNxiii  #1{\unskip}     \fi
\ifx \showISSN     \undefined \def \showISSN      #1{\unskip}     \fi
\ifx \showLCCN     \undefined \def \showLCCN      #1{\unskip}     \fi
\ifx \shownote     \undefined \def \shownote      #1{#1}          \fi
\ifx \showarticletitle \undefined \def \showarticletitle #1{#1}   \fi
\ifx \showURL      \undefined \def \showURL       {\relax}        \fi
\providecommand\bibfield[2]{#2}
\providecommand\bibinfo[2]{#2}
\providecommand\natexlab[1]{#1}
\providecommand\showeprint[2][]{arXiv:#2}

\bibitem[\protect\citeauthoryear{Ahmed, Le~Roux, Norouzi, and Schuurmans}{Ahmed
  et~al\mbox{.}}{2019}]%
        {ahmed2019understanding}
\bibfield{author}{\bibinfo{person}{Zafarali Ahmed}, \bibinfo{person}{Nicolas
  Le~Roux}, \bibinfo{person}{Mohammad Norouzi}, {and} \bibinfo{person}{Dale
  Schuurmans}.} \bibinfo{year}{2019}\natexlab{}.
\newblock \showarticletitle{Understanding the impact of entropy on policy
  optimization}. In \bibinfo{booktitle}{\emph{International Conference on
  Machine Learning}}. \bibinfo{pages}{151--160}.
\newblock


\bibitem[\protect\citeauthoryear{Ancona, Oztireli, and Gross}{Ancona
  et~al\mbox{.}}{2019}]%
        {ancona2019explaining}
\bibfield{author}{\bibinfo{person}{Marco Ancona}, \bibinfo{person}{Cengiz
  Oztireli}, {and} \bibinfo{person}{Markus Gross}.}
  \bibinfo{year}{2019}\natexlab{}.
\newblock \showarticletitle{Explaining deep neural networks with a polynomial
  time algorithm for shapley value approximation}. In
  \bibinfo{booktitle}{\emph{International Conference on Machine Learning}}.
  \bibinfo{pages}{272--281}.
\newblock


\bibitem[\protect\citeauthoryear{Bernstein, Givan, Immerman, and
  Zilberstein}{Bernstein et~al\mbox{.}}{2002}]%
        {bernstein2002complexity}
\bibfield{author}{\bibinfo{person}{Daniel~S Bernstein}, \bibinfo{person}{Robert
  Givan}, \bibinfo{person}{Neil Immerman}, {and} \bibinfo{person}{Shlomo
  Zilberstein}.} \bibinfo{year}{2002}\natexlab{}.
\newblock \showarticletitle{The complexity of decentralized control of Markov
  decision processes}.
\newblock \bibinfo{journal}{\emph{Mathematics of operations research}}
  \bibinfo{volume}{27}, \bibinfo{number}{4} (\bibinfo{year}{2002}),
  \bibinfo{pages}{819--840}.
\newblock


\bibitem[\protect\citeauthoryear{Bilbao and Edelman}{Bilbao and
  Edelman}{2000}]%
        {bilbao2000shapley}
\bibfield{author}{\bibinfo{person}{Jesus~Mario Bilbao} {and}
  \bibinfo{person}{Paul~H. Edelman}.} \bibinfo{year}{2000}\natexlab{}.
\newblock \showarticletitle{The Shapley value on convex geometries}.
\newblock \bibinfo{journal}{\emph{Discrete Applied Mathematics}}
  \bibinfo{volume}{103}, \bibinfo{number}{1-3} (\bibinfo{year}{2000}),
  \bibinfo{pages}{33--40}.
\newblock


\bibitem[\protect\citeauthoryear{Busoniu, Babuska, and De~Schutter}{Busoniu
  et~al\mbox{.}}{2008}]%
        {busoniu2008comprehensive}
\bibfield{author}{\bibinfo{person}{Lucian Busoniu}, \bibinfo{person}{Robert
  Babuska}, {and} \bibinfo{person}{Bart De~Schutter}.}
  \bibinfo{year}{2008}\natexlab{}.
\newblock \showarticletitle{A comprehensive survey of multiagent reinforcement
  learning}.
\newblock \bibinfo{journal}{\emph{IEEE Transactions on Systems, Man, and
  Cybernetics, Part C (Applications and Reviews)}} \bibinfo{volume}{38},
  \bibinfo{number}{2} (\bibinfo{year}{2008}), \bibinfo{pages}{156--172}.
\newblock


\bibitem[\protect\citeauthoryear{Cao, Yu, Ren, and Chen}{Cao
  et~al\mbox{.}}{2012}]%
        {cao2012overview}
\bibfield{author}{\bibinfo{person}{Yongcan Cao}, \bibinfo{person}{Wenwu Yu},
  \bibinfo{person}{Wei Ren}, {and} \bibinfo{person}{Guanrong Chen}.}
  \bibinfo{year}{2012}\natexlab{}.
\newblock \showarticletitle{An overview of recent progress in the study of
  distributed multi-agent coordination}.
\newblock \bibinfo{journal}{\emph{IEEE Transactions on Industrial informatics}}
  \bibinfo{volume}{9}, \bibinfo{number}{1} (\bibinfo{year}{2012}),
  \bibinfo{pages}{427--438}.
\newblock


\bibitem[\protect\citeauthoryear{Chen, Song, Wainwright, and Jordan}{Chen
  et~al\mbox{.}}{2018}]%
        {chen2018shapley}
\bibfield{author}{\bibinfo{person}{Jianbo Chen}, \bibinfo{person}{Le Song},
  \bibinfo{person}{Martin~J Wainwright}, {and} \bibinfo{person}{Michael~I
  Jordan}.} \bibinfo{year}{2018}\natexlab{}.
\newblock \showarticletitle{L-shapley and C-shapley: Efficient model
  interpretation for structured data}. In
  \bibinfo{booktitle}{\emph{International Conference on Learning
  Representations}}.
\newblock


\bibitem[\protect\citeauthoryear{Chen, Zhang, Xiao, He, Pu, and Chang}{Chen
  et~al\mbox{.}}{2019}]%
        {chen2019counterfactual}
\bibfield{author}{\bibinfo{person}{Long Chen}, \bibinfo{person}{Hanwang Zhang},
  \bibinfo{person}{Jun Xiao}, \bibinfo{person}{Xiangnan He},
  \bibinfo{person}{Shiliang Pu}, {and} \bibinfo{person}{Shih-Fu Chang}.}
  \bibinfo{year}{2019}\natexlab{}.
\newblock \showarticletitle{Counterfactual critic multi-agent training for
  scene graph generation}. In \bibinfo{booktitle}{\emph{Proceedings of the
  IEEE/CVF International Conference on Computer Vision}}.
  \bibinfo{pages}{4613--4623}.
\newblock


\bibitem[\protect\citeauthoryear{Fatima, Wooldridge, and Jennings}{Fatima
  et~al\mbox{.}}{2008}]%
        {fatima2008linear}
\bibfield{author}{\bibinfo{person}{Shaheen~S Fatima}, \bibinfo{person}{Michael
  Wooldridge}, {and} \bibinfo{person}{Nicholas~R Jennings}.}
  \bibinfo{year}{2008}\natexlab{}.
\newblock \showarticletitle{A linear approximation method for the Shapley
  value}.
\newblock \bibinfo{journal}{\emph{Artificial Intelligence}}
  \bibinfo{volume}{172}, \bibinfo{number}{14} (\bibinfo{year}{2008}),
  \bibinfo{pages}{1673--1699}.
\newblock


\bibitem[\protect\citeauthoryear{Foerster, Farquhar, Afouras, Nardelli, and
  Whiteson}{Foerster et~al\mbox{.}}{2018}]%
        {foerster2018counterfactual}
\bibfield{author}{\bibinfo{person}{Jakob Foerster}, \bibinfo{person}{Gregory
  Farquhar}, \bibinfo{person}{Triantafyllos Afouras}, \bibinfo{person}{Nantas
  Nardelli}, {and} \bibinfo{person}{Shimon Whiteson}.}
  \bibinfo{year}{2018}\natexlab{}.
\newblock \showarticletitle{Counterfactual multi-agent policy gradients}. In
  \bibinfo{booktitle}{\emph{Proceedings of the AAAI Conference on Artificial
  Intelligence}}, Vol.~\bibinfo{volume}{32}.
\newblock


\bibitem[\protect\citeauthoryear{Frye, Feige, and Rowat}{Frye
  et~al\mbox{.}}{2020}]%
        {frye2020asymmetric}
\bibfield{author}{\bibinfo{person}{Christopher Frye}, \bibinfo{person}{Ilya
  Feige}, {and} \bibinfo{person}{Colin Rowat}.}
  \bibinfo{year}{2020}\natexlab{}.
\newblock \showarticletitle{Asymmetric Shapley values: incorporating causal
  knowledge into model-agnostic explainability}.
\newblock \bibinfo{journal}{\emph{Conference and Workshop on Neural Information
  Processing Systems}}.
\newblock


\bibitem[\protect\citeauthoryear{Ghorbani and Zou}{Ghorbani and Zou}{2019}]%
        {ghorbani2019data}
\bibfield{author}{\bibinfo{person}{Amirata Ghorbani} {and}
  \bibinfo{person}{James Zou}.} \bibinfo{year}{2019}\natexlab{}.
\newblock \showarticletitle{Data shapley: Equitable valuation of data for
  machine learning}. In \bibinfo{booktitle}{\emph{International Conference on
  Machine Learning}}. \bibinfo{pages}{2242--2251}.
\newblock


\bibitem[\protect\citeauthoryear{Ghorbani and Zou}{Ghorbani and Zou}{2020}]%
        {ghorbani2020neuron}
\bibfield{author}{\bibinfo{person}{Amirata Ghorbani} {and}
  \bibinfo{person}{James Zou}.} \bibinfo{year}{2020}\natexlab{}.
\newblock \showarticletitle{Neuron shapley: Discovering the responsible
  neurons}.
\newblock \bibinfo{journal}{\emph{arXiv preprint arXiv:2002.09815}}
  (\bibinfo{year}{2020}).
\newblock


\bibitem[\protect\citeauthoryear{Gupta, Egorov, and Kochenderfer}{Gupta
  et~al\mbox{.}}{2017}]%
        {gupta2017cooperative}
\bibfield{author}{\bibinfo{person}{Jayesh~K Gupta}, \bibinfo{person}{Maxim
  Egorov}, {and} \bibinfo{person}{Mykel Kochenderfer}.}
  \bibinfo{year}{2017}\natexlab{}.
\newblock \showarticletitle{Cooperative multi-agent control using deep
  reinforcement learning}. In \bibinfo{booktitle}{\emph{International
  Conference on Autonomous Agents and Multiagent Systems}}.
  \bibinfo{pages}{66--83}.
\newblock


\bibitem[\protect\citeauthoryear{Heskes, Sijben, Bucur, and Claassen}{Heskes
  et~al\mbox{.}}{2020}]%
        {heskes2020causal}
\bibfield{author}{\bibinfo{person}{Tom Heskes}, \bibinfo{person}{Evi Sijben},
  \bibinfo{person}{Ioan~Gabriel Bucur}, {and} \bibinfo{person}{Tom Claassen}.}
  \bibinfo{year}{2020}\natexlab{}.
\newblock \showarticletitle{Causal Shapley Values: Exploiting Causal Knowledge
  to Explain Individual Predictions of Complex Models}.
\newblock \bibinfo{journal}{\emph{Conference and Workshop on Neural Information
  Processing Systems}}.
\newblock


\bibitem[\protect\citeauthoryear{Heuillet, Couthouis, and
  D{\'\i}az-Rodr{\'\i}guez}{Heuillet et~al\mbox{.}}{2021}]%
        {heuillet2021explainability}
\bibfield{author}{\bibinfo{person}{Alexandre Heuillet}, \bibinfo{person}{Fabien
  Couthouis}, {and} \bibinfo{person}{Natalia D{\'\i}az-Rodr{\'\i}guez}.}
  \bibinfo{year}{2021}\natexlab{}.
\newblock \showarticletitle{Explainability in deep reinforcement learning}.
\newblock \bibinfo{journal}{\emph{Knowledge-Based Systems}}
  \bibinfo{volume}{214} (\bibinfo{year}{2021}), \bibinfo{pages}{106685}.
\newblock


\bibitem[\protect\citeauthoryear{Iqbal and Sha}{Iqbal and Sha}{2019}]%
        {iqbal2019actor}
\bibfield{author}{\bibinfo{person}{Shariq Iqbal} {and} \bibinfo{person}{Fei
  Sha}.} \bibinfo{year}{2019}\natexlab{}.
\newblock \showarticletitle{Actor-attention-critic for multi-agent
  reinforcement learning}. In \bibinfo{booktitle}{\emph{International
  Conference on Machine Learning}}. \bibinfo{pages}{2961--2970}.
\newblock


\bibitem[\protect\citeauthoryear{Keviczky, Borrelli, Fregene, Godbole, and
  Balas}{Keviczky et~al\mbox{.}}{2007}]%
        {keviczky2007decentralized}
\bibfield{author}{\bibinfo{person}{Tam{\'a}s Keviczky},
  \bibinfo{person}{Francesco Borrelli}, \bibinfo{person}{Kingsley Fregene},
  \bibinfo{person}{Datta Godbole}, {and} \bibinfo{person}{Gary~J Balas}.}
  \bibinfo{year}{2007}\natexlab{}.
\newblock \showarticletitle{Decentralized receding horizon control and
  coordination of autonomous vehicle formations}.
\newblock \bibinfo{journal}{\emph{IEEE Transactions on control systems
  technology}} \bibinfo{volume}{16}, \bibinfo{number}{1}
  (\bibinfo{year}{2007}), \bibinfo{pages}{19--33}.
\newblock


\bibitem[\protect\citeauthoryear{Kraemer and Banerjee}{Kraemer and
  Banerjee}{2016}]%
        {kraemer2016multi}
\bibfield{author}{\bibinfo{person}{Landon Kraemer} {and}
  \bibinfo{person}{Bikramjit Banerjee}.} \bibinfo{year}{2016}\natexlab{}.
\newblock \showarticletitle{Multi-agent reinforcement learning as a rehearsal
  for decentralized planning}.
\newblock \bibinfo{journal}{\emph{Neurocomputing}}  \bibinfo{volume}{190}
  (\bibinfo{year}{2016}), \bibinfo{pages}{82--94}.
\newblock


\bibitem[\protect\citeauthoryear{Kumar, Venkatasubramanian, Scheidegger, and
  Friedler}{Kumar et~al\mbox{.}}{2020}]%
        {kumar2020problems}
\bibfield{author}{\bibinfo{person}{I~Elizabeth Kumar}, \bibinfo{person}{Suresh
  Venkatasubramanian}, \bibinfo{person}{Carlos Scheidegger}, {and}
  \bibinfo{person}{Sorelle Friedler}.} \bibinfo{year}{2020}\natexlab{}.
\newblock \showarticletitle{Problems with Shapley-value-based explanations as
  feature importance measures}. In \bibinfo{booktitle}{\emph{International
  Conference on Machine Learning}}. \bibinfo{pages}{5491--5500}.
\newblock


\bibitem[\protect\citeauthoryear{Lillicrap, Hunt, Pritzel, Heess, Erez, Tassa,
  Silver, and Wierstra}{Lillicrap et~al\mbox{.}}{2016}]%
        {lillicrap2016continuous}
\bibfield{author}{\bibinfo{person}{Timothy~P Lillicrap},
  \bibinfo{person}{Jonathan~J Hunt}, \bibinfo{person}{Alexander Pritzel},
  \bibinfo{person}{Nicolas Heess}, \bibinfo{person}{Tom Erez},
  \bibinfo{person}{Yuval Tassa}, \bibinfo{person}{David Silver}, {and}
  \bibinfo{person}{Daan Wierstra}.} \bibinfo{year}{2016}\natexlab{}.
\newblock \showarticletitle{Continuous control with deep reinforcement
  learning}.
\newblock \bibinfo{journal}{\emph{International Conference on Learning
  Representations}} (\bibinfo{year}{2016}).
\newblock


\bibitem[\protect\citeauthoryear{Lowe, Wu, Tamar, Harb, Abbeel, and
  Mordatch}{Lowe et~al\mbox{.}}{2017}]%
        {lowe2017multi}
\bibfield{author}{\bibinfo{person}{Ryan Lowe}, \bibinfo{person}{Yi Wu},
  \bibinfo{person}{Aviv Tamar}, \bibinfo{person}{Jean Harb},
  \bibinfo{person}{Pieter Abbeel}, {and} \bibinfo{person}{Igor Mordatch}.}
  \bibinfo{year}{2017}\natexlab{}.
\newblock \showarticletitle{Multi-agent actor-critic for mixed
  cooperative-competitive environments}. In
  \bibinfo{booktitle}{\emph{Proceedings of the 31st International Conference on
  Neural Information Processing Systems}}. \bibinfo{pages}{6382--6393}.
\newblock


\bibitem[\protect\citeauthoryear{Meng}{Meng}{2012}]%
        {meng2012core}
\bibfield{author}{\bibinfo{person}{Fan-Yong Meng}.}
  \bibinfo{year}{2012}\natexlab{}.
\newblock \showarticletitle{The Core and Shapley Function for Games on
  Augmenting Systems with a Coalition Structure}.
\newblock \bibinfo{journal}{\emph{International Journal of Mathematical and
  Computational Sciences}} \bibinfo{volume}{6}, \bibinfo{number}{8}
  (\bibinfo{year}{2012}), \bibinfo{pages}{813--818}.
\newblock


\bibitem[\protect\citeauthoryear{Mnih, Badia, Mirza, Graves, Lillicrap, Harley,
  Silver, and Kavukcuoglu}{Mnih et~al\mbox{.}}{2016}]%
        {mnih2016asynchronous}
\bibfield{author}{\bibinfo{person}{Volodymyr Mnih},
  \bibinfo{person}{Adria~Puigdomenech Badia}, \bibinfo{person}{Mehdi Mirza},
  \bibinfo{person}{Alex Graves}, \bibinfo{person}{Timothy Lillicrap},
  \bibinfo{person}{Tim Harley}, \bibinfo{person}{David Silver}, {and}
  \bibinfo{person}{Koray Kavukcuoglu}.} \bibinfo{year}{2016}\natexlab{}.
\newblock \showarticletitle{Asynchronous methods for deep reinforcement
  learning}. In \bibinfo{booktitle}{\emph{International conference on machine
  learning}}. \bibinfo{pages}{1928--1937}.
\newblock


\bibitem[\protect\citeauthoryear{Oliehoek and Amato}{Oliehoek and
  Amato}{2016}]%
        {oliehoek2016concise}
\bibfield{author}{\bibinfo{person}{Frans~A Oliehoek} {and}
  \bibinfo{person}{Christopher Amato}.} \bibinfo{year}{2016}\natexlab{}.
\newblock \bibinfo{booktitle}{\emph{A concise introduction to decentralized
  POMDPs}}.
\newblock \bibinfo{publisher}{Springer}.
\newblock


\bibitem[\protect\citeauthoryear{Oliehoek, Spaan, and Vlassis}{Oliehoek
  et~al\mbox{.}}{2008}]%
        {oliehoek2008optimal}
\bibfield{author}{\bibinfo{person}{Frans~A Oliehoek},
  \bibinfo{person}{Matthijs~TJ Spaan}, {and} \bibinfo{person}{Nikos Vlassis}.}
  \bibinfo{year}{2008}\natexlab{}.
\newblock \showarticletitle{Optimal and approximate Q-value functions for
  decentralized POMDPs}.
\newblock \bibinfo{journal}{\emph{Journal of Artificial Intelligence Research}}
   \bibinfo{volume}{32} (\bibinfo{year}{2008}), \bibinfo{pages}{289--353}.
\newblock


\bibitem[\protect\citeauthoryear{Palmer, Tuyls, Bloembergen, and Savani}{Palmer
  et~al\mbox{.}}{2018}]%
        {palmer2018lenient}
\bibfield{author}{\bibinfo{person}{Gregory Palmer}, \bibinfo{person}{Karl
  Tuyls}, \bibinfo{person}{Daan Bloembergen}, {and} \bibinfo{person}{Rahul
  Savani}.} \bibinfo{year}{2018}\natexlab{}.
\newblock \showarticletitle{Lenient Multi-Agent Deep Reinforcement Learning}.
  In \bibinfo{booktitle}{\emph{Proceedings of the 17th International Conference
  on Autonomous Agents and MultiAgent Systems}}. \bibinfo{pages}{443--451}.
\newblock


\bibitem[\protect\citeauthoryear{Ramchurn, Farinelli, Macarthur, and
  Jennings}{Ramchurn et~al\mbox{.}}{2010}]%
        {ramchurn2010decentralized}
\bibfield{author}{\bibinfo{person}{Sarvapali~D Ramchurn},
  \bibinfo{person}{Alessandro Farinelli}, \bibinfo{person}{Kathryn~S
  Macarthur}, {and} \bibinfo{person}{Nicholas~R Jennings}.}
  \bibinfo{year}{2010}\natexlab{}.
\newblock \showarticletitle{Decentralized coordination in robocup rescue}.
\newblock \bibinfo{journal}{\emph{Comput. J.}} \bibinfo{volume}{53},
  \bibinfo{number}{9} (\bibinfo{year}{2010}), \bibinfo{pages}{1447--1461}.
\newblock


\bibitem[\protect\citeauthoryear{Rashid, Samvelyan, Schroeder, Farquhar,
  Foerster, and Whiteson}{Rashid et~al\mbox{.}}{2018}]%
        {rashid2018qmix}
\bibfield{author}{\bibinfo{person}{Tabish Rashid}, \bibinfo{person}{Mikayel
  Samvelyan}, \bibinfo{person}{Christian Schroeder}, \bibinfo{person}{Gregory
  Farquhar}, \bibinfo{person}{Jakob Foerster}, {and} \bibinfo{person}{Shimon
  Whiteson}.} \bibinfo{year}{2018}\natexlab{}.
\newblock \showarticletitle{Qmix: Monotonic value function factorisation for
  deep multi-agent reinforcement learning}. In
  \bibinfo{booktitle}{\emph{International Conference on Machine Learning}}.
  \bibinfo{pages}{4295--4304}.
\newblock


\bibitem[\protect\citeauthoryear{Samvelyan, Rashid, De~Witt, Farquhar,
  Nardelli, Rudner, Hung, Torr, Foerster, and Whiteson}{Samvelyan
  et~al\mbox{.}}{2019}]%
        {samvelyan2019starcraft}
\bibfield{author}{\bibinfo{person}{Mikayel Samvelyan}, \bibinfo{person}{Tabish
  Rashid}, \bibinfo{person}{Christian~Schroeder De~Witt},
  \bibinfo{person}{Gregory Farquhar}, \bibinfo{person}{Nantas Nardelli},
  \bibinfo{person}{Tim~GJ Rudner}, \bibinfo{person}{Chia-Man Hung},
  \bibinfo{person}{Philip~HS Torr}, \bibinfo{person}{Jakob Foerster}, {and}
  \bibinfo{person}{Shimon Whiteson}.} \bibinfo{year}{2019}\natexlab{}.
\newblock \showarticletitle{The starcraft multi-agent challenge}.
\newblock  (\bibinfo{year}{2019}).
\newblock


\bibitem[\protect\citeauthoryear{Shapley}{Shapley}{1953}]%
        {shapley1953value}
\bibfield{author}{\bibinfo{person}{Lloyd~S Shapley}.}
  \bibinfo{year}{1953}\natexlab{}.
\newblock \showarticletitle{A value for n-person games}.
\newblock \bibinfo{journal}{\emph{Contributions to the Theory of Games}}
  (\bibinfo{year}{1953}).
\newblock


\bibitem[\protect\citeauthoryear{Silver, Lever, Heess, Degris, Wierstra, and
  Riedmiller}{Silver et~al\mbox{.}}{2014}]%
        {silver2014deterministic}
\bibfield{author}{\bibinfo{person}{David Silver}, \bibinfo{person}{Guy Lever},
  \bibinfo{person}{Nicolas Heess}, \bibinfo{person}{Thomas Degris},
  \bibinfo{person}{Daan Wierstra}, {and} \bibinfo{person}{Martin Riedmiller}.}
  \bibinfo{year}{2014}\natexlab{}.
\newblock \showarticletitle{Deterministic policy gradient algorithms}. In
  \bibinfo{booktitle}{\emph{International conference on machine learning}}.
  \bibinfo{pages}{387--395}.
\newblock


\bibitem[\protect\citeauthoryear{Son, Kim, Kang, Hostallero, and Yi}{Son
  et~al\mbox{.}}{2019}]%
        {son2019qtran}
\bibfield{author}{\bibinfo{person}{Kyunghwan Son}, \bibinfo{person}{Daewoo
  Kim}, \bibinfo{person}{Wan~Ju Kang}, \bibinfo{person}{David~Earl Hostallero},
  {and} \bibinfo{person}{Yung Yi}.} \bibinfo{year}{2019}\natexlab{}.
\newblock \showarticletitle{Qtran: Learning to factorize with transformation
  for cooperative multi-agent reinforcement learning}. In
  \bibinfo{booktitle}{\emph{International Conference on Machine Learning}}.
  \bibinfo{pages}{5887--5896}.
\newblock


\bibitem[\protect\citeauthoryear{Sundararajan and Najmi}{Sundararajan and
  Najmi}{2020}]%
        {sundararajan2020many}
\bibfield{author}{\bibinfo{person}{Mukund Sundararajan} {and}
  \bibinfo{person}{Amir Najmi}.} \bibinfo{year}{2020}\natexlab{}.
\newblock \showarticletitle{The many Shapley values for model explanation}. In
  \bibinfo{booktitle}{\emph{International Conference on Machine Learning}}.
  \bibinfo{pages}{9269--9278}.
\newblock


\bibitem[\protect\citeauthoryear{Sunehag, Lever, Gruslys, Czarnecki, Zambaldi,
  Jaderberg, Lanctot, Sonnerat, Leibo, Tuyls, et~al\mbox{.}}{Sunehag
  et~al\mbox{.}}{2018}]%
        {sunehag2018value}
\bibfield{author}{\bibinfo{person}{Peter Sunehag}, \bibinfo{person}{Guy Lever},
  \bibinfo{person}{Audrunas Gruslys}, \bibinfo{person}{Wojciech~Marian
  Czarnecki}, \bibinfo{person}{Vinicius Zambaldi}, \bibinfo{person}{Max
  Jaderberg}, \bibinfo{person}{Marc Lanctot}, \bibinfo{person}{Nicolas
  Sonnerat}, \bibinfo{person}{Joel~Z Leibo}, \bibinfo{person}{Karl Tuyls},
  {et~al\mbox{.}}} \bibinfo{year}{2018}\natexlab{}.
\newblock \showarticletitle{Value-Decomposition Networks For Cooperative
  Multi-Agent Learning Based On Team Reward}. In
  \bibinfo{booktitle}{\emph{Proceedings of the 17th International Conference on
  Autonomous Agents and MultiAgent Systems}}. \bibinfo{pages}{2085--2087}.
\newblock


\bibitem[\protect\citeauthoryear{Tan}{Tan}{1993}]%
        {tan1993multi}
\bibfield{author}{\bibinfo{person}{Ming Tan}.} \bibinfo{year}{1993}\natexlab{}.
\newblock \showarticletitle{Multi-agent reinforcement learning: Independent vs.
  cooperative agents}. In \bibinfo{booktitle}{\emph{Proceedings of the tenth
  international conference on machine learning}}. \bibinfo{pages}{330--337}.
\newblock


\bibitem[\protect\citeauthoryear{Wang, Wiens, and Lundberg}{Wang
  et~al\mbox{.}}{2021}]%
        {wang2021shapley}
\bibfield{author}{\bibinfo{person}{Jiaxuan Wang}, \bibinfo{person}{Jenna
  Wiens}, {and} \bibinfo{person}{Scott Lundberg}.}
  \bibinfo{year}{2021}\natexlab{}.
\newblock \showarticletitle{Shapley Flow: A Graph-based Approach to
  Interpreting Model Predictions}.
\newblock  (\bibinfo{year}{2021}).
\newblock


\bibitem[\protect\citeauthoryear{Wang, Zhang, Kim, and Gu}{Wang
  et~al\mbox{.}}{2020}]%
        {wang2020shapleyq}
\bibfield{author}{\bibinfo{person}{Jianhong Wang}, \bibinfo{person}{Yuan
  Zhang}, \bibinfo{person}{Tae-Kyun Kim}, {and} \bibinfo{person}{Yunjie Gu}.}
  \bibinfo{year}{2020}\natexlab{}.
\newblock \showarticletitle{Shapley Q-value: A Local Reward Approach to Solve
  Global Reward Games}. In \bibinfo{booktitle}{\emph{Proceedings of the AAAI
  Conference on Artificial Intelligence}}, Vol.~\bibinfo{volume}{34}.
  \bibinfo{pages}{7285--7292}.
\newblock


\bibitem[\protect\citeauthoryear{Williams and Peng}{Williams and Peng}{1991}]%
        {williams1991function}
\bibfield{author}{\bibinfo{person}{Ronald~J Williams} {and}
  \bibinfo{person}{Jing Peng}.} \bibinfo{year}{1991}\natexlab{}.
\newblock \showarticletitle{Function optimization using connectionist
  reinforcement learning algorithms}.
\newblock \bibinfo{journal}{\emph{Connection Science}} \bibinfo{volume}{3},
  \bibinfo{number}{3} (\bibinfo{year}{1991}), \bibinfo{pages}{241--268}.
\newblock


\bibitem[\protect\citeauthoryear{Wolpert and Tumer}{Wolpert and Tumer}{2002}]%
        {wolpert2002optimal}
\bibfield{author}{\bibinfo{person}{David~H Wolpert} {and}
  \bibinfo{person}{Kagan Tumer}.} \bibinfo{year}{2002}\natexlab{}.
\newblock \showarticletitle{Optimal payoff functions for members of
  collectives}.
\newblock In \bibinfo{booktitle}{\emph{Modeling complexity in economic and
  social systems}}. \bibinfo{publisher}{World Scientific},
  \bibinfo{pages}{355--369}.
\newblock


\bibitem[\protect\citeauthoryear{Yang, Hao, Chen, Tang, Chen, Hu, Fan, and
  Wei}{Yang et~al\mbox{.}}{2020a}]%
        {yang2020q}
\bibfield{author}{\bibinfo{person}{Yaodong Yang}, \bibinfo{person}{Jianye Hao},
  \bibinfo{person}{Guangyong Chen}, \bibinfo{person}{Hongyao Tang},
  \bibinfo{person}{Yingfeng Chen}, \bibinfo{person}{Yujing Hu},
  \bibinfo{person}{Changjie Fan}, {and} \bibinfo{person}{Zhongyu Wei}.}
  \bibinfo{year}{2020}\natexlab{a}.
\newblock \showarticletitle{Q-value path decomposition for deep multiagent
  reinforcement learning}. In \bibinfo{booktitle}{\emph{International
  Conference on Machine Learning}}. \bibinfo{pages}{10706--10715}.
\newblock


\bibitem[\protect\citeauthoryear{Yang, Hao, Liao, Shao, Chen, Liu, and
  Tang}{Yang et~al\mbox{.}}{2020b}]%
        {yang2020qatten}
\bibfield{author}{\bibinfo{person}{Yaodong Yang}, \bibinfo{person}{Jianye Hao},
  \bibinfo{person}{Ben Liao}, \bibinfo{person}{Kun Shao},
  \bibinfo{person}{Guangyong Chen}, \bibinfo{person}{Wulong Liu}, {and}
  \bibinfo{person}{Hongyao Tang}.} \bibinfo{year}{2020}\natexlab{b}.
\newblock \showarticletitle{Qatten: A General Framework for Cooperative
  Multiagent Reinforcement Learning}.
\newblock \bibinfo{journal}{\emph{arXiv e-prints}} (\bibinfo{year}{2020}).
\newblock


\bibitem[\protect\citeauthoryear{Yang, Luo, Li, Zhou, Zhang, and Wang}{Yang
  et~al\mbox{.}}{2018}]%
        {yang2018mean}
\bibfield{author}{\bibinfo{person}{Yaodong Yang}, \bibinfo{person}{Rui Luo},
  \bibinfo{person}{Minne Li}, \bibinfo{person}{Ming Zhou},
  \bibinfo{person}{Weinan Zhang}, {and} \bibinfo{person}{Jun Wang}.}
  \bibinfo{year}{2018}\natexlab{}.
\newblock \showarticletitle{Mean field multi-agent reinforcement learning}. In
  \bibinfo{booktitle}{\emph{International Conference on Machine Learning}}.
  \bibinfo{pages}{5571--5580}.
\newblock


\bibitem[\protect\citeauthoryear{Ye, Zhang, and Yang}{Ye et~al\mbox{.}}{2015}]%
        {ye2015multi}
\bibfield{author}{\bibinfo{person}{Dayong Ye}, \bibinfo{person}{Minjie Zhang},
  {and} \bibinfo{person}{Yun Yang}.} \bibinfo{year}{2015}\natexlab{}.
\newblock \showarticletitle{A multi-agent framework for packet routing in
  wireless sensor networks}.
\newblock \bibinfo{journal}{\emph{sensors}} \bibinfo{volume}{15},
  \bibinfo{number}{5} (\bibinfo{year}{2015}), \bibinfo{pages}{10026--10047}.
\newblock


\bibitem[\protect\citeauthoryear{Zhou, Liu, Sui, Li, and Chung}{Zhou
  et~al\mbox{.}}{2020}]%
        {zhou2020learning}
\bibfield{author}{\bibinfo{person}{Meng Zhou}, \bibinfo{person}{Ziyu Liu},
  \bibinfo{person}{Pengwei Sui}, \bibinfo{person}{Yixuan Li}, {and}
  \bibinfo{person}{Yuk~Ying Chung}.} \bibinfo{year}{2020}\natexlab{}.
\newblock \showarticletitle{Learning Implicit Credit Assignment for Multi-Agent
  Actor-Critic}.
\newblock \bibinfo{journal}{\emph{Conference and Workshop on Neural Information
  Processing Systems}}.
\newblock


\end{thebibliography}

\end{document}